\documentclass[final]{cvpr}

\usepackage{times}
\usepackage{epsfig}
\usepackage{graphicx}
\usepackage{amsmath}
\usepackage{amssymb}

\usepackage{subfigure}
\usepackage{booktabs} 
\usepackage{bbm}
\usepackage{pifont}
\usepackage{nopageno}

\usepackage{algorithm}
\usepackage[noend]{algpseudocode}

\usepackage{algorithmicx}

\newcommand{\cmark}{\ding{51}}%
\newcommand{\xmark}{\ding{55}}%

\newcommand{\MS}[1]{{\color{blue}  #1}}
\newcommand{\YSR}[1]{{\color{blue} {\bf YSR:} #1}}

\usepackage[pagebackref=true,breaklinks=true,colorlinks,bookmarks=false]{hyperref}

\def\cvprPaperID{65} 
\def\confYear{CVPR 2021}
\begin{document}

\title{PLM: Partial Label Masking for Imbalanced Multi-label Classification}

\author{Kevin Duarte \\ 
\tt \small kevin\_duarte@knights.ucf.edu \\
\and 
Yogesh Rawat \\ 
\tt \small yogesh@crcv.ucf.edu\\ 
\and  
Mubarak Shah \\ 
\tt \small shah@crcv.ucf.edu\\
\and
Center for Research in Computer Vision\\
University of Central Florida, Orlando, Florida, USA\\
}

\maketitle

\begin{abstract}
Neural networks trained on real-world datasets with long-tailed label distributions are biased towards frequent classes and perform poorly on infrequent classes. The imbalance in the ratio of positive and negative samples for each class skews network output probabilities further from ground-truth distributions. We propose a method, Partial Label Masking (PLM), which utilizes this ratio during training. By stochastically masking labels during loss computation, the method balances this ratio for each class, leading to improved recall on minority classes and improved precision on frequent classes. The ratio is estimated adaptively based on the network's performance by minimizing the KL divergence between predicted and ground-truth distributions. 
Whereas most existing approaches addressing data imbalance are mainly focused on \textit{single-label} classification and do not generalize well to the multi-label case, this work proposes a general approach to solve the long-tail data imbalance issue for \textit{multi-label} classification. PLM is versatile: it can be applied to most objective functions and it can be used alongside other strategies for class imbalance. Our method achieves strong performance when compared to existing methods on both \textit{multi-label} (MultiMNIST and MSCOCO) and \textit{single-label} (imbalanced CIFAR-10 and CIFAR-100) image classification datasets.

\end{abstract}


\section{Introduction}

The impressive performance of deep learning methods has led to the creation of many large-scale datasets \cite{deng2009imagenet,gu2018ava, lin2014microsoft,openimages}. Due to the naturally imbalanced distribution of objects within the world, these datasets contain imbalanced numbers of samples for different classes. The class labels in these datasets form a long-tailed distribution: several classes appear frequently (the head classes), while many classes contain few samples (the tail classes). This imbalance causes classifiers to perform poorly, especially on classes which are infrequent in training. Finding a solution to this problem is necessary to successfully scale deep networks to larger real-world datasets which tend to have long-tail data distributions. 

\begin{figure}[t]

\begin{center}
   \includegraphics[width=0.49\linewidth]{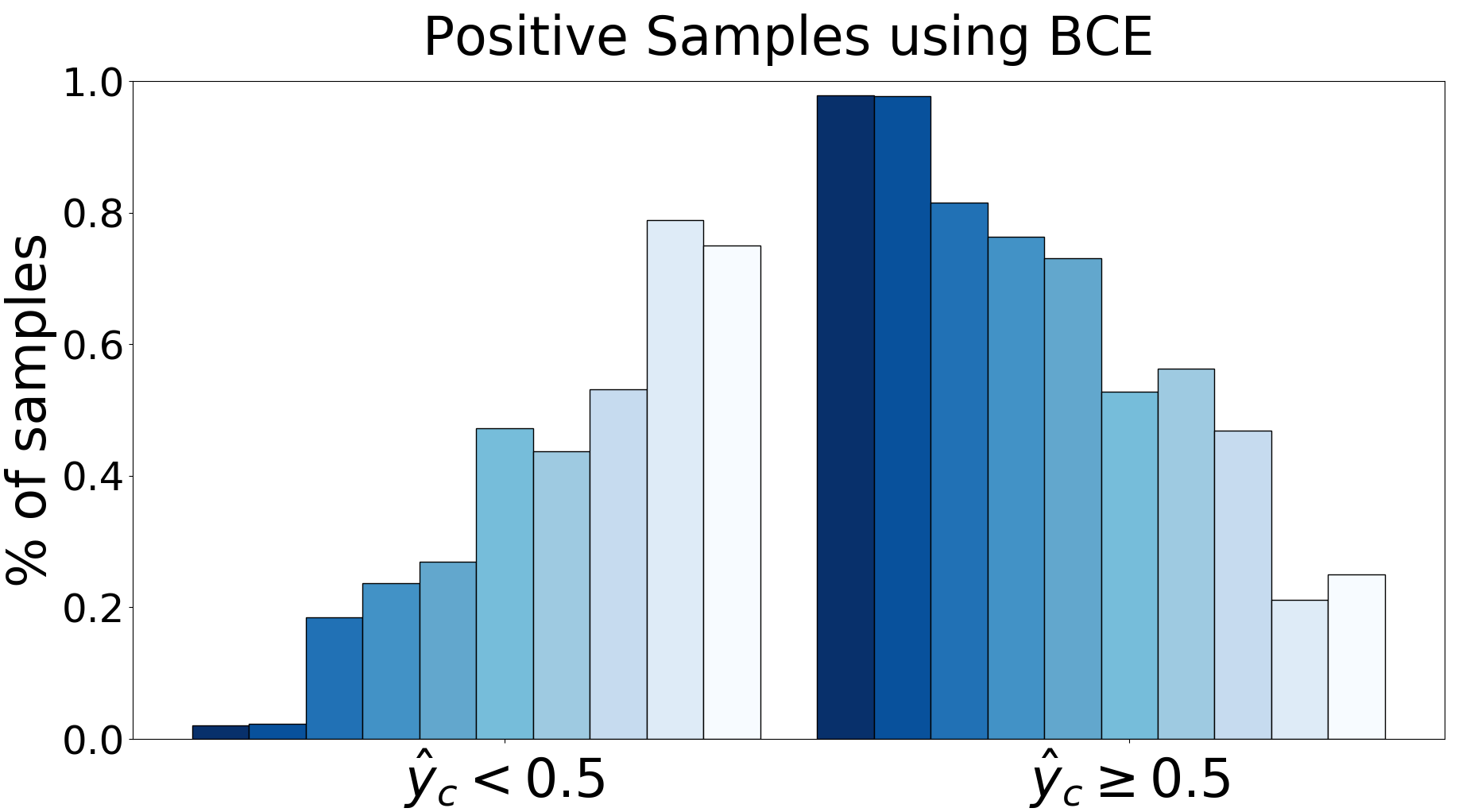}
   \includegraphics[width=0.49\linewidth]{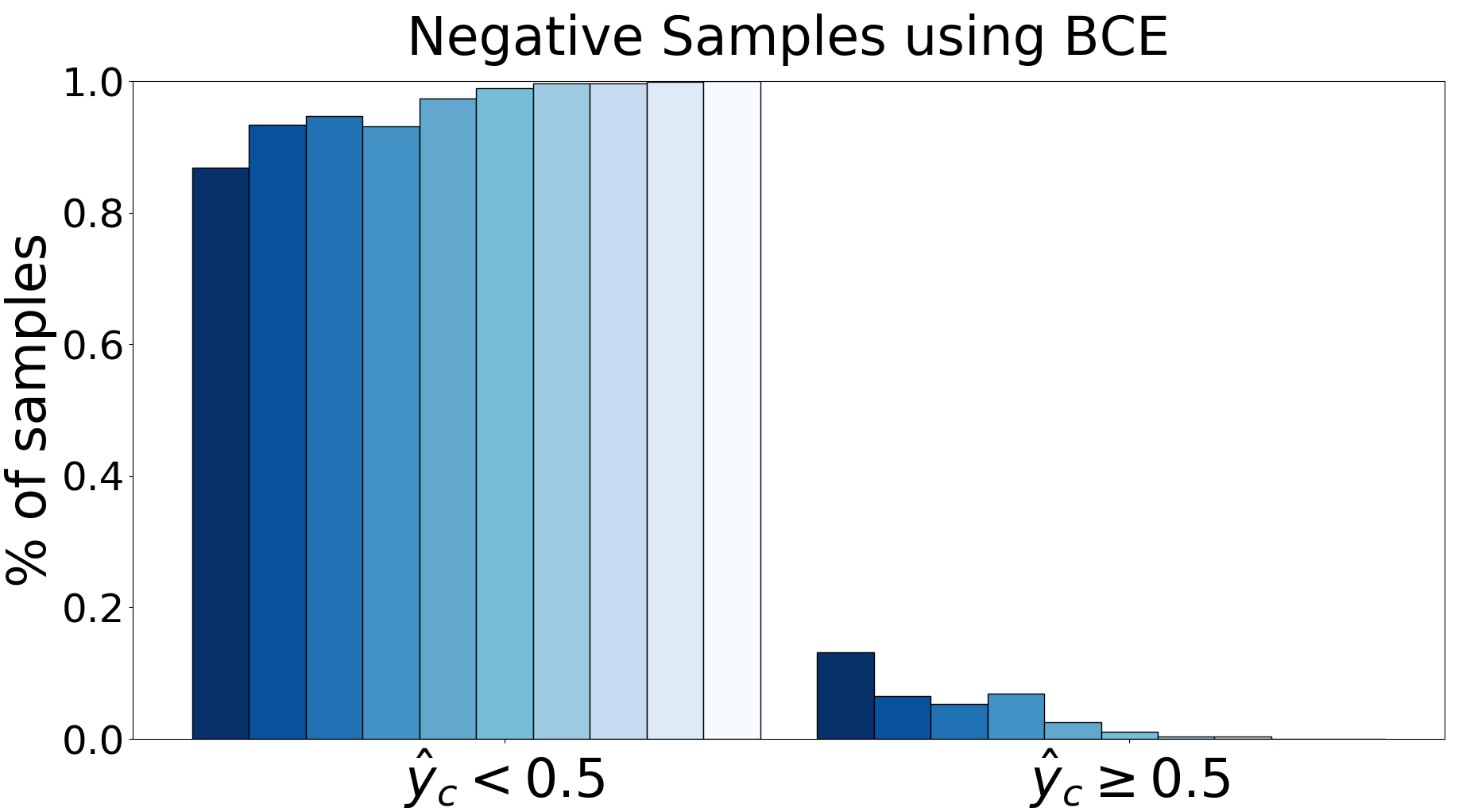}
   \includegraphics[width=0.49\linewidth]{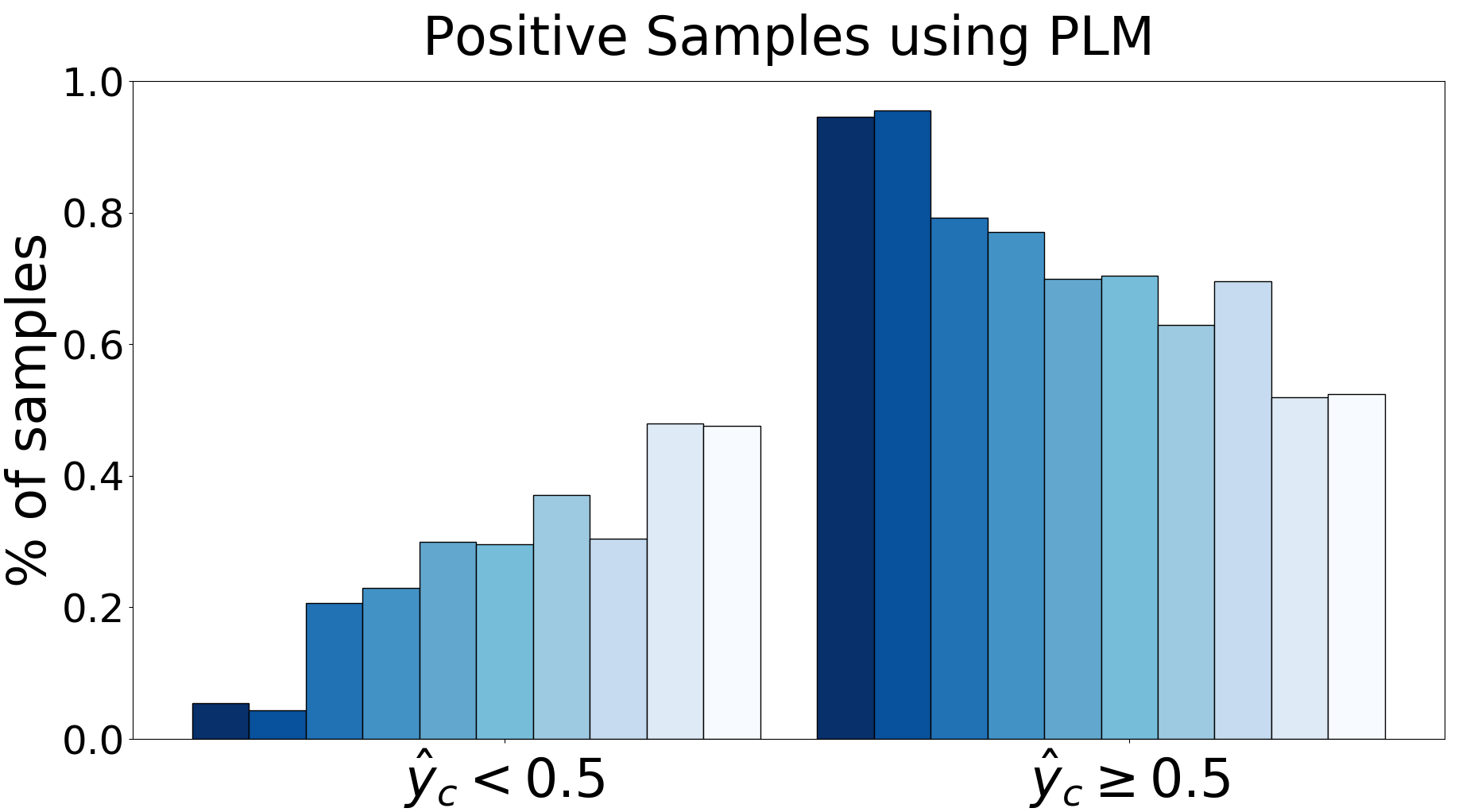}
   \includegraphics[width=0.49\linewidth]{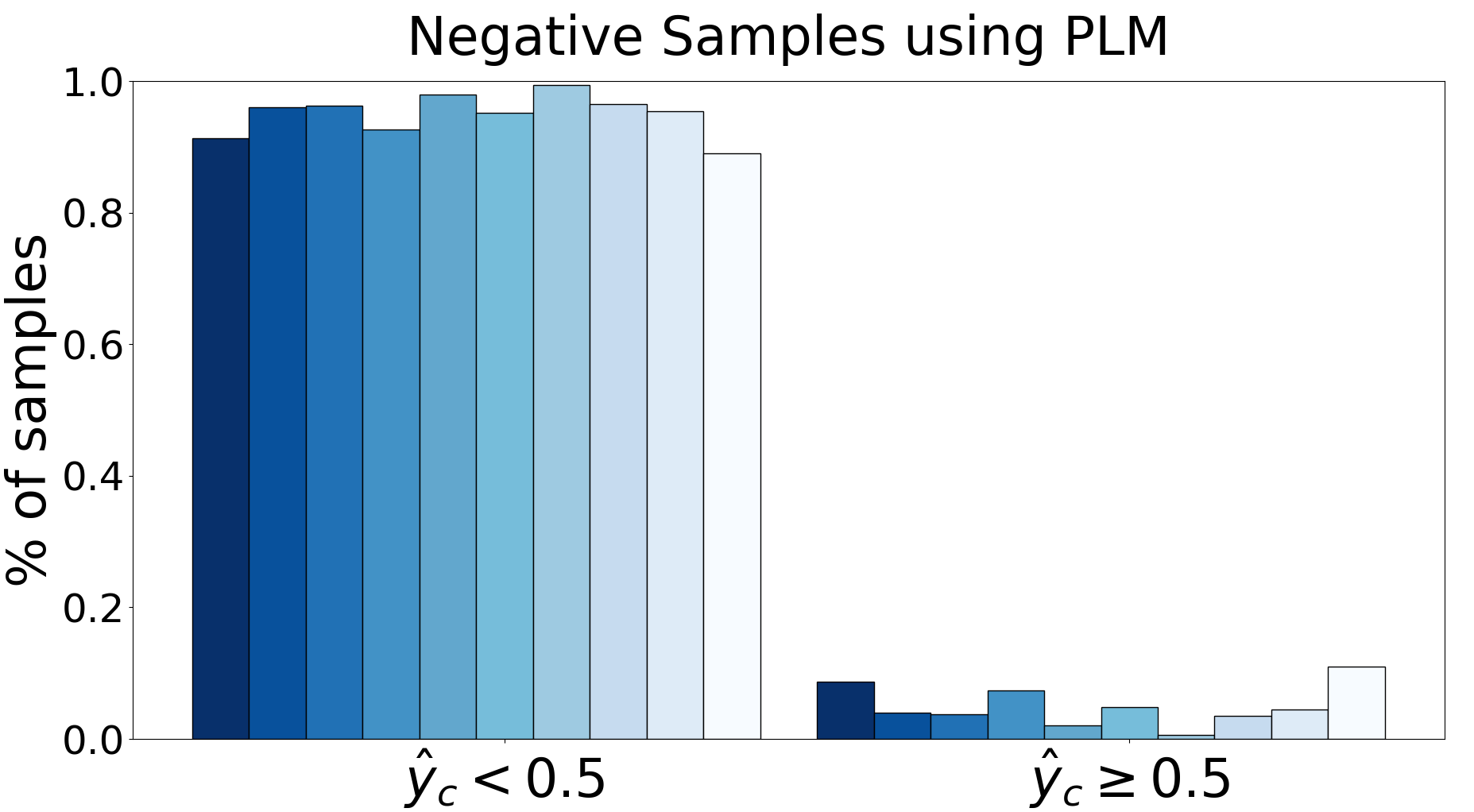}
   \includegraphics[width=\linewidth]{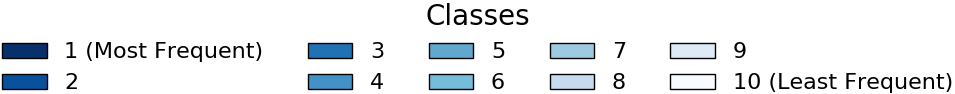}
\end{center}
\vspace{-0.2cm}
   \caption{The output probability ($\hat{y}_c$) distributions of a ResNet-32 classifier trained on artificially imbalanced CIFAR-10 for positive (left) and negative samples (right). For frequent classes, the predicted distribution skews towards 1; for infrequent classes, it skews towards 0. Classifiers trained using PLM (bottom) reduces this bias, when compared to classifiers trained with binary cross-entropy (top).}
\vspace{-0.3cm}
\label{fig:distributionsintro}
\end{figure}

Several recent works \cite{buda2018systematic,cui2019class,cao2019learning} attempt to solve the data imbalance issues; however, most tend to have single-label assumptions. For example, LDAM-DRW \cite{cao2019learning} performs very well in single-label settings, but it assumes a single class label, $y$, is present for a given sample, $x$, to compute class margins $\gamma (x, y)$ for their proposed loss. Not only that, but the two most common methods for learning long-tailed distributions, reweighting and resampling, were not designed for data with multiple labels. Our experiments show that reweighting based on the inverse number of samples performs poorly on multi-label datasets; also, it is difficult to resample multi-label data due to the co-occurrence of labels within individual samples. Since many real-world applications like image tagging \cite{liu2017semantic,gattupalli2019weakly}, recommendation systems \cite{zhang2019deep,lu2019learning}, and action detection \cite{gu2018ava,feichtenhofer2019slowfast}, often involve multi-label classification and suffer from imbalanced data, we believe that class-imbalance methods should be developed for both single-label and multi-label settings. Therefore, in this work, we propose a general solution for long-tailed imbalance which works for both multi-label and single-label classification.

Classifiers trained on imbalanced multi-label datasets tend to over-predict frequent classes and under-predict minority classes. This behaviour is displayed in Figure \ref{fig:distributionsintro}. When the class is not present within the image, the probability output for the frequent classes is skewed towards 1; conversely, when an infrequent class is present within the sample, the classifier outputs a low probability score. These output probability distributions differ greatly from the ideal distribution (i.e. the ground-truth distribution where all positive samples are labeled 1 and all negative samples are labeled 0). {\em We argue, that this behaviour is caused  not only by an imbalance in the number of positive samples between different classes, but also by the ratio of positive and negative samples for each class}.

As dataset imbalance increases, the ratio of positive samples to negative samples increases for head classes and decreases for tail classes. We find that the change in this ratio greatly impacts the classifier's ability to generalize. If a class has large ratio of positive samples to negative samples, the classifier over-predicts the given class, leading to an increase of false positive predictions; conversely, a small ratio leads to under-prediction and an increase of false negatives. Assuming there is an optimal ratio which can minimize the over/under-predictions, an algorithm can estimate and leverage this ratio to improve network performance. 

We present Partial Label Masking (PLM): a novel approach for training classifiers on imbalanced multi-label datasets which improves network generalization by leveraging this ratio. 
By partially masking positive and negative labels for frequent and infrequent classes respectively; our method reduces the discrepancy between the classifiers' output probability distribution and the ground-truth distribution (as seen in Figure \ref{fig:distributionsintro}). Our method performs this masking stochastically for each sample and it continually adapts the target ratio based on the classifier's output probabilities. This leads to improved precision on classes with many samples and improved recall on classes with few samples. Moreover, our method consistently improves performance on difficult classes, regardless of the number of samples.


Our contributions include: (i) we present a general solution for data imbalance which balances the ratio between positive and negative samples, (ii) we propose an adaptive strategy to determine the ideal ratio which minimizes the difference between predicted probability and ground-truth  distributions, (iii) we empirically evaluate our method on both multi-label datasets (imbalanced MultiMNIST and MSCOCO) and single-label datasets 
(CIFAR10 and CIFAR100), and (iv) we thoroughly analyse our method's ability to improve classifiers' performance on both difficult and infrequent classes. 


\section{Method}

\paragraph{Notation} We denote a dataset with $N$ samples and $C$ class categories as $\mathcal{D}= \{\left( x^{\left(i\right)}, \textbf{y}^{\left(i\right)}\right)\}$ where $x^{\left(i\right)}$ is the $i^{th}$ input and $\textbf{y}^{\left(i\right)} = [ y^{\left(i\right)}_1,...,y^{\left(i\right)}_C]$ is its corresponding binary label vector; note that multiple elements in this label may be non-zero in multi-label classification. The number of positive samples for a class $c$ is $n_c^+ = |\{i : i \in \{1,...,N\} \textnormal{, where } y^{\left(i\right)}_c = 1\}|$, and the number of negative samples is $n_c^-=N-n_c^+$. The ratio of positive samples to negative samples for a class $c$ is $r_c=\frac{n_c^+}{n_c^-}$. We denote the ``optimal" ratio, which minimizes over/under-predictions as $\bar{r}_c$.

\subsection{Partial Label Masking}

Given a network prediction, $\hat{\textbf{y}}^{\left(i\right)}$, the total loss is sum of the losses over all $C$ classes:
\begin{equation}
\label{eq:loss}
    L\left(\textbf{y}^{\left(i\right)}, \hat{\textbf{y}}^{\left(i\right)}\right) = \sum_{j=1}^C \ell \left(y^{\left(i\right)}_j, \hat{y}^{\left(i\right)}_j\right).
\end{equation}
This is a general form (e.g. for binary cross-entropy, $\ell \left(y, \hat{y}\right) = -\left[y\log \left( \hat{y} \right) + \left(1-y\right)\log \left(1-\hat{y}\right)\right]$).

Partial Label Masking (PLM) masks the loss computed for certain classes. For each image $i$ we generate a binary mask $\textbf{g}^{\left(i\right)} = [ g^{\left(i\right)}_1,...,g^{\left(i\right)}_C]$ that masks loss calculations for certain classes as follows: 
\begin{equation}
\label{eq:plm}
    L\left(\textbf{y}^{\left(i\right)}, \hat{\textbf{y}}^{\left(i\right)}, \textbf{g}^{\left(i\right)}\right) = \sum_{j=1}^C g^{\left(i\right)}_j \ell \left(y^{\left(i\right)}_j, \hat{y}^{\left(i\right)}_j\right).
\end{equation}

Note that equation \ref{eq:plm} is a generalized form of \ref{eq:loss} when all masks equal one, i.e. $g^{\left(i\right)}_c=1$. {\em Therefore, PLM can be used with any loss which sums over different classes.} 

\paragraph{Masks Generation} With this formulation, one can generate masks so that frequent class samples  are ignored by the loss calculation and an equal number of samples for each class is used to train the network, which is typically employed in under-sampling. 
The issue with under-sampling samples, however, is that frequent and infrequent classes can co-occur within the same sample, so removing it from training would not meaningfully change the imbalance. The benefit of the PLM formulation is that individual positive, or negative, labels can be masked to ensure a specific number of labels for each class is used to train the network regardless of co-occurrence.


Instead of under-sampling based on the number of positive samples, we consider the ratio of positive to negative samples for each class, $r_c$. Assuming there is some ideal ratio, $\bar{r}_c$, which minimizes over/under-predictions, we can generate the masks, $\textbf{g}^{\left(i\right)}$, stochastically as a function of the label:
\begin{equation}
\label{eq:cases}
    g^{\left(i\right)}_c = f\left(y^{\left(i\right)}_c\right)=
    \begin{cases}
        \mathbbm{1}\left[\frac{\bar{r}_c}{r_c}\right], & r_c > \bar{r}_c \text{ and }  y^{\left(i\right)}_c = 1\\
        \mathbbm{1}\left[\frac{r_c}{\bar{r}_c}\right], & r_c < \bar{r}_c \text{ and }  y^{\left(i\right)}_c = 0\\
        1, &\text{otherwise}
    \end{cases}
\end{equation}
where $\mathbbm{1}\left[p\right]= 1$ with probability $p$, and 0 with probability $1-p$. The probabilities used, ${\bar{r}_c}/{r_c}$ and ${r_c}/{\bar{r}_c}$, ensure that the ratio between positive and negative samples on which loss is calculated is $\bar{r}_c$. With partial label masking, we are able to train a classifier with a set ratio between positives and negatives for each class. Now the question arises: How does one determine this ideal ratio?

\paragraph{Ratio Selection}
The ratio between positives and negatives for each class should be set to minimize over-predictions and under-predictions for the head and tail classes respectively. Since decreasing the ratio reduces the number of positive samples used for loss calculation and increasing the ratio reduces the number of negative samples, the ratio selected for the head classes should be decreased (i.e. $\bar{r}_c < r_c$). On the other hand, the ratio for the tail classes should be increased ($\bar{r}_c > r_c$). 

One simple approach for setting the ratio would be to use the average ratio of all the classes. This ensures a decrease in the ratio for head classes and increase for tail classes. Although, this ratio leads to improved results on some datasets, it is sensitive to different levels of imbalance within the dataset as well as the relative difficulty of the various classes. Ideally this hyper-parameter would change based on the performance of the classifier on the dataset on-hand. To this end, we propose a method to estimate the ratio adaptively based on the output probability distribution of the classifier.

\paragraph{Ratio Adaptation } For a class $c$, the set of output probabilities for the positive and negative samples are represented by $\hat{S}^+_c$ and $\hat{S}^-_c$, respectively. The ground-truth sets are $S^+_c$ and $S^-_c$, where members of this set will be 1 for a positive sample and 0 for a negative sample. Formally, these sets can be defined as:
\begin{equation}
\begin{aligned}
  S^+_c = \{y_c^{\left(i\right)} | y_c^{\left(i\right)}=1, \forall i\}, \\
  \hat{S}^+_c = \{\hat{y}_c^{\left(i\right)} | y_c^{\left(i\right)}=1, \forall i\}, \\
  S^-_c = \{y_c^{\left(i\right)} | y_c^{\left(i\right)}=0, \forall i\}, \\
  \hat{S}^-_c = \{\hat{y}_c^{\left(i\right)} | y_c^{\left(i\right)}=0, \forall i\}. \\
\end{aligned}
\end{equation}
A discrete distribution\footnote{Additional information on how this distribution is formed can be found in the Appendix.} is formed using these sets by placing the probabilities in $\tau$ bins of width $1/\tau$. For each set, we  denote these distributions as $P^+_c$, $\hat{P}^+_c$, $P^-_c$, and $\hat{P}^-_c$ respectively. When data imbalance is present, the classifiers output probabilities skew further from the ground-truth distributions. For head classes, the difference between $\hat{P}^-_c$ and $P^-_c$ increases as the probability outputs for negative samples become larger; conversely, for tail classes, probabilities are pushed closer to $0$ for positive samples, so the difference between $\hat{P}^+_c$ and $P^+_c$ increases.

\begin{algorithm*}[t]
    \caption{PLM training algorithm. Inputs are a network $F_\theta$ and the dataset $\mathcal{D}=\{\left( x^{\left(i\right)}, \textbf{y}^{\left(i\right)}\right)\}_{i=1}^N$. }
    \label{alg:plm}
\begin{algorithmic}[1]
\State Initialize $\bar{r}_{c,1}$  

\For{$t=1$ \textbf{to} $T$ epochs} 
  \State Generate masks $g_c^{\left(i\right)} = f\left(y^{\left(i\right)}_c\right)$, using $\bar{r}_{c,t}$ \Comment{Equation \ref{eq:cases}}
  \Repeat
      \State $\mathcal{B} \leftarrow \text{SampleMiniBatch}\left(\mathcal{D}\right)$ 
      \State $\hat{\textbf{y}} \leftarrow F_\theta\left(\mathcal{B}\right)$ \Comment{Perform forward pass on mini-batch}
      \State $F_\theta \leftarrow F_\theta - \alpha \nabla_\theta L\left(\textbf{y}, \hat{\textbf{y}}, \textbf{g}\right)$ \Comment{One SGD step with loss in equation \ref{eq:plm}}
  \Until all data has been sampled
  \State Generate $P^+_c, P^-_c, \hat{P}^+_c, \hat{P}^-_c$ from all $\textbf{y}^{\left(i\right)}$ and $\hat{\textbf{y}}^{\left(i\right)}$ 
  \State Compute $D_c =\tilde{D}^+_c - \tilde{D}^-_c$ using equations \ref{eq:dists} and \ref{eq:distsnorm}
  \State $\bar{r}_{c,t+1} \leftarrow e^{\lambda D_c }\bar{r}_{c,t}$ \Comment{Update ratio}
\EndFor
\end{algorithmic}
\end{algorithm*}

We utilize this difference to change the ratio until the output distributions better resemble the ground-truth distribution. To compute the difference between the distributions, we use the Kullback-Leibler divergence: 
\begin{equation}
\label{eq:dists}
    D^+_c = D_\text{KL} \left(\hat{P}^+_c || P^+_c\right);\text{ and } D^-_c = D_\text{KL} \left(\hat{P}^-_c || P^-_c \right).
\end{equation}
We normalize the divergence
\begin{equation}
\label{eq:distsnorm}
    \tilde{D}^+_c = \frac{D^+_c - \mu^+}{\sigma^+};\text{ and } \tilde{D}^-_c = \frac{D^-_c - \mu^-}{\sigma^-},
\end{equation}
with their means ($\mu^+$ and $\mu^-$) and standard deviations ($\sigma^+$ and $\sigma^-$). After normalization, $\tilde{D}^+_c$ tends to be positive when the network under-predicts class $c$ and negative when the network over-predicts. The inverse holds for $\tilde{D}^-_c$: it is positive when the network over-predicts, and negative otherwise.

The ratio should be adjusted during training until the divergence scores in equation \ref{eq:distsnorm} become balanced ($D^+_c=D^-_c$). At each epoch $t$, the ratio $\bar{r}_{c,t}$ becomes
\begin{equation}
\label{eq:ratiochange}
    \bar{r}_{c,t} = e^{\lambda D_c }\bar{r}_{c,t-1},
\end{equation}
where $\bar{r}_{c,1}$ is the initial ratio (e.g. the dataset's ratio, $r_c$), $D_c = \tilde{D}^+_c - \tilde{D}^-_c$, and $\lambda$ is a hyper-parameter which controls the rate of change of the ratio. The exponential term in equation \ref{eq:ratiochange} increases, or decreases, the current ratio for each class based on the divergence between ground-truth and output probability distributions: if $D_c>0$, which tends to occur for infrequent classes, then the ratio increases; if $D_c<0$, which occurs for frequent classes, then the ratio decreases. If $D_c=0$ (i.e. $D^+_c=D^-_c$), then the ratio remains unchanged as some optimal ratio has been reached. The training procedure used for partial label masking is described in Algorithm \ref{alg:plm}.

\section{Experimental Evaluation}

We evaluate our proposed method on several image classification benchmarks with varying levels of imbalance. Imbalance, $\rho$,  is denoted as the ratio between the number of samples for the most frequent class and the least frequent class: $\rho = \max_i\{n^+_i\}/\min_i\{n^+_i\}$.

\subsection{Multi-label Classification}

\paragraph{Datasets} We evaluate our method on two multi-label datasets: MultiMNIST and the large-scale real-world multi-label dataset MSCOCO \cite{lin2014microsoft}. We 
construct the imbalanced MultiMNIST dataset by superimposing two MNIST \cite{lecun1998mnist} digits into a single image. We sample the MNIST training set to obtain a long-tail distribution over the different digits. Due to coocurrence of the same digit within a sample, the final imbalance of the dataset is $\rho=90.33$. The test set consists of 90000 samples; it is roughly balanced, with all classes being present a similar number of samples\footnote{Additional details of the dataset's construction and the dataset statistics are in the Appendix.}. MSCOCO is a multi-label image classification benchmark, which contain a large amount of imbalance ($\rho=352.92$). We use the standard train/test split which consists of 82,783 training and 40,504 evaluation images. It contains 80 classes, with an average of 2.9 labels per sample.
\vspace{-0.2cm}

\paragraph{Metrics}
We evaluate classifier performance using multiple standard multi-label classification metrics: per-class precision, per-class recall, F1-score, and 0-1 exact match accuracy. To evaluate performance on tail classes, we measure some metrics averaged over the $K$ most infrequent classes. A description of these metrics is in the Appendix.
\vspace{-0.2cm}

\paragraph{Baselines} Since recent class imbalance works tend to focus on single-label classification, we compare our method with those approaches which can be applied to multi-label classification. We use the following baselines: i) binary cross-entropy, ii) focal loss \cite{lin2017focal}, iii) reweighting by inverse number of classes (CB), and iv) undersampling. For reweighting, we use the formulation:
\begin{equation}
    L\left(\textbf{y}^{\left(i\right)}, \hat{\textbf{y}}^{\left(i\right)}\right) = \sum_{j=1}^C w_j \ell \left(y^{\left(i\right)}_j, \hat{y}^{\left(i\right)}_j\right),
\end{equation}
where $w_j$ is the inverse number of samples for class $j$, as calculated in \cite{cui2019class}. Undersampling that removes all imbalance is a non-trivial operation when multiple labels can are present in each sample. Therefore, for each epoch, we select $S$ samples from each class (without duplication) to reduce the imbalance seen by the network (S=500 for MultiMNIST and S=1000 for MSCOCO).

\begin{table*}[t!]
    \centering
    \begin{tabular}{l|c|c|ccc|ccc|c}
        \hline
        Loss  & Und. & Precision & \multicolumn{3}{c|}{Recall} & \multicolumn{3}{c|}{F1 Score} & 0-1 Acc. \\
         & & Avg. & Avg. & K=5 & K=3 & Avg. & K=5 & K=3 &   \\
        \hline
        BCE & \xmark & 86.96 & 78.65 & 63.39 & 51.50 & 80.35 & 73.87 & 65.73 &  53.42  \\
        Focal  & \xmark & 86.75 & 78.28 & 62.53 & 50.44 & 79.93 & 73.33 & 64.88 &  52.64   \\
        BCE+CB & \xmark & 85.86  & 76.55 & 61.02 & 49.12 & 78.05 & 72.32 & 63.81 &  48.51  \\
        BCE & \cmark & 87.97 & 79.00 & 65.36 & 54.16 & 81.39 & 75.57 & 67.92 &  54.46  \\
        \hline
        PLM (BCE) & \xmark &  87.41 & \textbf{81.55} & 71.91 & 65.31 & \textbf{83.73} & 79.17 & 74.27 & \textbf{56.95}   \\
        PLM (Focal) & \xmark  & 87.50 & 80.95 & 70.27 & 62.31 & 83.22 & 78.13 & 72.61 & 56.45     \\
        PLM (BCE+CB) & \xmark & \textbf{89.37} & 75.24 & 65.31 & 56.25 & 80.52 & 75.64 & 69.28 & 49.95   \\
        PLM (BCE) & \cmark & 88.54 & 80.38 & \textbf{71.29} & \textbf{65.86} & 83.56 & \textbf{79.44} & \textbf{75.10} & 56.53 \\
        \hline
    \end{tabular}
    \caption{Results for multi-label image classification on MultiMNIST. ``Und." denotes undersampling and ``CB" denotes class-based reweighting \cite{cui2019class}.
    }
    \label{tab:mnistresults}
\end{table*}

\begin{table*}[t!]
    \centering
    \begin{tabular}{l|c|c|ccc|ccc|c}
        \hline
        Loss &  Und. & Precision & \multicolumn{3}{c|}{Recall} & \multicolumn{3}{c|}{F1 Score} & 0-1 Acc. \\
         & & Avg. & Avg. & K=20 & K=10 & Avg. & K=20 & K=10 &   \\
        \hline
        BCE & \xmark & \textbf{69.35} & 39.89 & 30.02 & 16.00 & 48.74 & 37.94 & 21.80 & \textbf{21.76}  \\
        Focal & \xmark & 68.01 & 38.62 & 28.92 & 15.10 & 47.38 & 36.52 & 20.46 &  20.88  \\
        BCE+CB & \xmark & 67.82  & 37.23 & 35.25 & 23.68 & 46.10 & 43.64 & 30.53 &  19.43 \\
        BCE  & \cmark & 65.96  & 37.75 & 35.18 & 19.50  & 46.06 & 41.59 & 24.16 &  18.60  \\
        \hline
        PLM (BCE) & \xmark &  67.75 & \textbf{41.29} & 39.69 & 29.91 & \textbf{50.16} & 44.24 & 31.26 & 20.51  \\
        PLM (Focal) & \xmark & 66.89 & 39.48 & 37.63 & 27.67 & 48.49 & 42.38 & 29.56 & 19.66  \\
        PLM (BCE+CB) & \xmark & 65.71 & 37.06 & 39.06 & 29.25 & 46.39 & \textbf{45.68} & \textbf{33.18} & 17.41  \\
        PLM (BCE) & \cmark & 63.72 & 36.23 & \textbf{40.21} & \textbf{31.46} & 44.85 & 44.49 & 32.72 & 16.71 \\
        \hline
    \end{tabular}
    \caption{Results for multi-label image classification on MSCOCO. ``Und.", denotes undersampling and ``CB" denotes class-based reweighting \cite{cui2019class}.\vspace{-0.3cm}}
    \label{tab:mscocoresults}
\end{table*}

\paragraph{Implementation} For experiments on MultiMNIST, we train a ResNet-12 \cite{he2016deep} model using stochastic gradient descent (SGD) with momentum of 0.9 for 90 epochs. The initial learning rate is set to 0.1 which is decayed by a factor of 0.1 at epochs 60 and 80. We employ a linear learning rate warm-up \cite{goyal2017accurate} for the first 5 epochs. The network is model with a batch size of 128. On MSCOCO, a ResNet-50 model is trained with a similar training procedure; however, the initial learning rate is set to 0.4 which is decayed epochs 30, 60, and 80, and the batch size is increased to 200. For both datasets, we initialize the ratio with the dataset's ratio (i.e. $\bar{r}_{c,1}=r_c$) and set $\lambda$ to $0.01$. Since the imbalance in MSCOCO is large, we clip the ratios within the range $[0,1]$.

\paragraph{Results on MultiMNIST}
We present results for MultiMNIST in Table \ref{tab:mnistresults}. Undersampling improves results on MultiMNIST across all metrics, but the use of reweighting (CB) leads to a decrease in performance since the technique is formulated for the single-label case. The use of PLM during training improves results across all metrics. The F1-score improves by 2.34\% and the exact match accuracy improves by 2.49\% over the next best baseline. Notably, PLM greatly improves performance on the minority classes - for the three most infrequent digits, the recall increases by 11.7\% and the F1-score increases by 7.18\%.

\paragraph{Results on MSCOCO}
Results for MSCOCO are presented in Table \ref{tab:mscocoresults}. We find conventional methods designed for single-label data imbalance (focal loss, class-based reweighting, and under-sampling) tend to underperform standard binary cross-entropy (BCE) training. Although reweighting and under-sampling lead to improvements on tail classes, they dramatically reduce performance on head classes leading to an overall performance decrease. PLM, on the other hand, achieves an overall improvement on recall (1.4\%) and f1-score (1.42\%). There is a decrease in precision due to the drastic improvement of recall on minority classes - this is further discussed in section \ref{sect:analysis}. Whereas PLM leads to a 0-1 accuracy improvement on MultiMNIST, there is a decrease on MSCOCO due to the imbalance of the MSCOCO test set and the bias of this metric to frequently occurring classes. Overall, PLM greatly improves performance on tail classes: when compared to the BCE baseline, the 10 most infrequent classes have a 13.91\% increase in recall and a 9.46\% increase in f1-score. Moreover, these experiments highlight the versatility of our method: PLM can be used on top of existing data imbalance methods as well as different losses.

\begin{table*}[t!]
    \centering
    \begin{tabular}{l|cc|cc|cc|cc|cc}
        \hline
        Imbalance & \multicolumn{2}{c|}{200} & \multicolumn{2}{c|}{100} & \multicolumn{2}{c|}{50} & \multicolumn{2}{c|}{20} & \multicolumn{2}{c}{10}  \\
        & Overall & K=50 & Overall & K=50 & Overall & K=50 & Overall & K=50 & Overall & K=50 \\
        \hline
        BCE & 34.88 & 55.84 & 29.43 & 46.46 & 23.34 & 35.52 & 16.37 & 22.54 & 12.71 & 15.98\\
        Focal & 36.28 & 58.80 & 29.69 & 46.88 & 23.38 & 35.22 & 16.53 & 23.00  & 12.63 & 16.74\\
        CB \cite{cui2019class} & 31.11 & - & 25.43 & - & 20.73 & - & 15.64 & - & 12.51 & - \\
        LDAM-DRW \cite{cao2019learning} & - & - & \textbf{22.97} & - & - & - &  - & - & \textbf{11.84} & - \\
        \hline
        PLM (BCE) & \textbf{29.67} & 44.65 & 24.86 & \textbf{36.05} & 20.50 & 28.66 & \textbf{15.21} & 19.92 & 12.35 & 14.98  \\
        PLM (Focal) & 29.79 & \textbf{44.54} & 25.19 & 36.53 & \textbf{20.19} & \textbf{27.94} & \textbf{15.21} & \textbf{19.60} & 12.20 & \textbf{14.60}\\
        \hline
    \end{tabular}
    \caption{Error rates for single-label image classification on imbalanced CIFAR-10. We measure the average error across all classes (Overall) as well as the error on the 5 classes with the fewest number of samples (K=5).}
    \label{tab:cifar10results}
\end{table*}

\begin{table*}[t!]
    \centering
    \begin{tabular}{l|cc|cc|cc|cc|cc}
        \hline
        
        Imbalance & \multicolumn{2}{c|}{200} & \multicolumn{2}{c|}{100} & \multicolumn{2}{c|}{50} & \multicolumn{2}{c|}{20} & \multicolumn{2}{c}{10}  \\
         & Overall & K=50 & Overall & K=50 & Overall & K=50 & Overall & K=50 & Overall & K=50 \\
        \hline
        BCE & 63.92 & 86.10 & 59.88 & 79.18 & 55.25 & 72.84 & 48.66 & 59.98 & 43.35 & 52.12  \\
        Focal & 64.67 & 87.84 & 60.04 & 81.62 & 55.71 & 74.60 & 48.62 & 63.86 & 43.04 & 53.74  \\
        CB \cite{cui2019class} & 63.77 & - & 60.40 & - & 54.68 &-& 47.41 & - & 42.01 & - \\
        LDAM-DRW \cite{cao2019learning} & - &-  & 57.96 & - & - &- & - & - & \textbf{41.29} & - \\
        \hline
        PLM (BCE) & \textbf{61.37} & 80.79 & \textbf{55.98} & 73.14 & 52.49 & 67.07 & 46.15 & 57.26 & 41.71 & 49.78 \\
        PLM (Focal) & 61.68 & \textbf{79.43} & 56.19 & \textbf{72.39} & \textbf{52.09} & \textbf{65.39} & \textbf{46.11} & \textbf{56.28} & 41.54 & \textbf{48.83} \\
        \hline
    \end{tabular}
    \caption{Error Rates for single-label image classification on imbalanced CIFAR-100. We measure the average error across all classes (Overall) as well as the error on the 50 classes with the fewest number of samples (K=50). \vspace{-0.2cm}}
    \label{tab:cifar100results}
\end{table*}

\subsection{Single-label Classification}
\paragraph{Datasets}
CIFAR-10 and CIFAR-100 \cite{krizhevsky2009learning} contain $32\times32$ images with 10 and 100 classes respectively. To artificially create imbalance, we reduce the training examples for certain classes, while keeping the 10,000 validation images unchanged. Following the experimental setup in \cite{cui2019class}, we evaluate on varying levels of long-tailed imbalance, ranging from 10 to 200.

\paragraph{Baselines} We compare our method with the following baselines: i) binary-cross entropy (BCE), ii) focal loss, iii) reweighting (CB) \cite{cui2019class}, and iv) LDAM-DRW \cite{cao2019learning}. The later two are recent methods for dealing with imbalanced single-label datasets.

\paragraph{Implementation}
For the CIFAR experiments, we train a ResNet-32 following the training procedure described in  \cite{cui2019class}. Again, we initialize the ratio with the dataset's ratio. The hyper-parameter $\lambda$ is set to 1.0 and 0.01 for the CIFAR-10 and CIFAR-100 experiments respectively. These values were selected empirically - we analyse the effect of this hyper-parameter in section \ref{sect:ablations}.

\paragraph{Results}
Tables \ref{tab:cifar10results} and \ref{tab:cifar100results} contain PLM's results on CIFAR-10 and CIFAR-100, respecively, for several imbalance factors ($\rho=10,20,50,100,200$). PLM achieves improvement when compared to the common methods for data imbalance - focal loss and reweighting based on the inverse effective number of samples (CB). Although the LDAM-DRW approach tends to achieve lower error in these experiments, it is based on a single-label assumption (i.e. equations 10 and 12 in their work require only a single positive label to be be present within a sample) and cannot be applied to multi-label classification. On the other hand, our approach achieves strong performance in both single-label and multi-label settings. 

\subsection{Ablations}
\label{sect:ablations}
In this section, we run several experiments to understand the importance of the hyperparameter, $\lambda$, and the initial ratio, $\bar{r}_{c,1}$.

\begin{figure}[t]
\centering
\begin{center}
   \includegraphics[width=\linewidth]{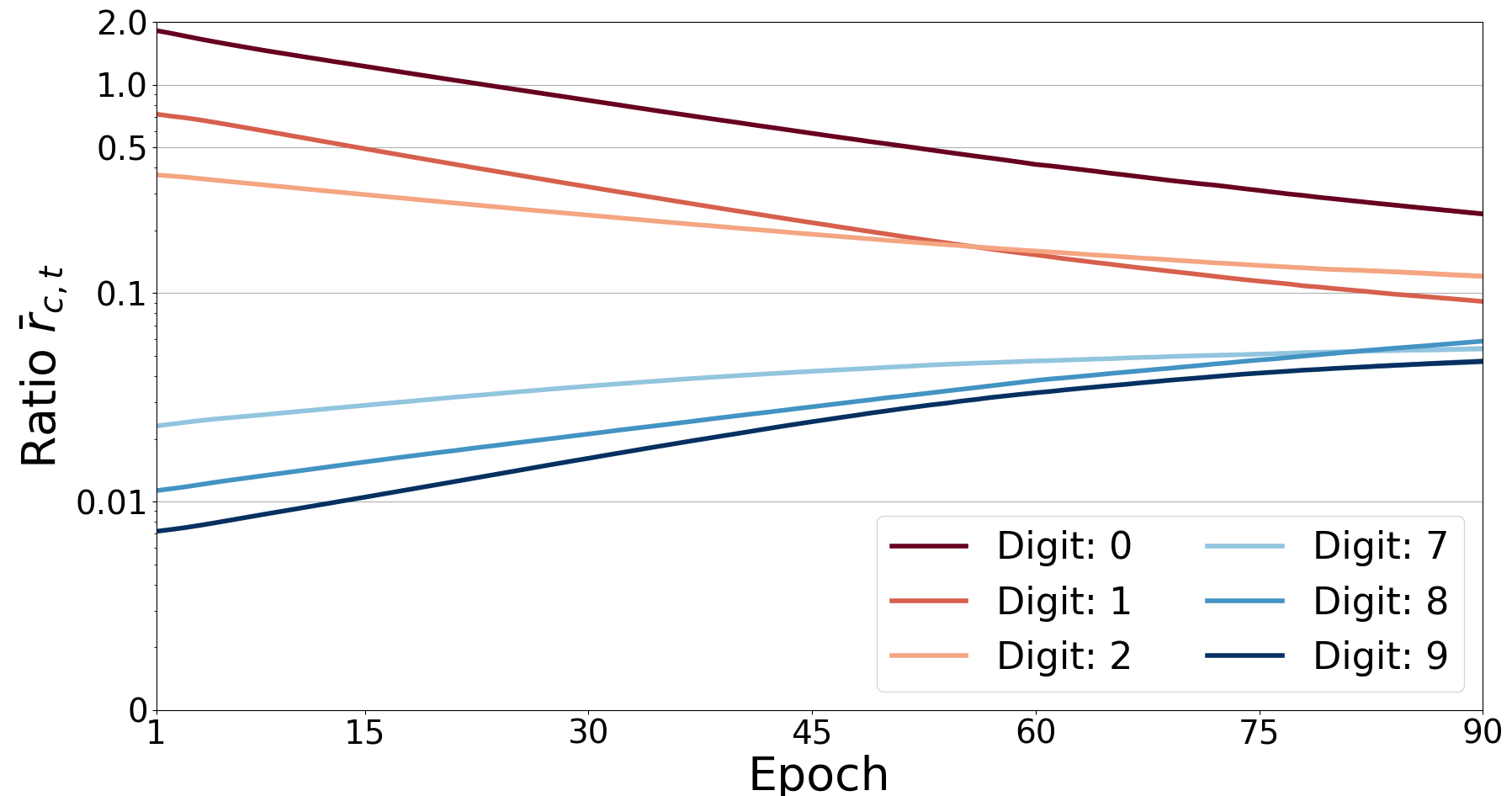}
\end{center}
\vspace{-0.2cm}
   \caption{The change of the ratio over time on imbalanced MultiMNIST. The ratio for tail classes tend to increase while the ratio for head classes tend to decrease.}
\label{fig:ratiochange}
\end{figure}

\paragraph{Ratio Adaptation} Since our approach attempts to estimate an optimal ratio for training a network, we investigate how the ratio changes throughout training. We present the effect of ratio's rate of change ($\lambda$) on the CIFAR-10 and CIFAR-100 datasets in Table \ref{tab:ablationlambda}. We find that all tested values lead to an improvement over no change (i.e. using standard BCE). On CIFAR-10, a higher rate of change ($\lambda=1.0$) leads to best performance. Meanwhile, a lower rate of change ($\lambda=0.01$) leads to best results on CIFAR-100 (this is also the case on MultiMNIST and MSCOCO). Figure \ref{fig:ratiochange} depicts how the ratio changes throughout training on the imbalanced MultiMNIST dataset. The ratio of head classes (digits 0, 1, and 2) decrease, while the ratio of the tail classes (digits 7, 8, and 9) increase. Over the 90 epochs, the ratio of the most frequent class (digit 0) decreases from 1.82 to 0.24 (decreasing by a factor of 7.58), and the ratio of the least frequent class (digit 9) increases from 0.007 to 0.047 (increasing by a factor of 6.54). This demonstrates that the original ratios (i.e. the dataset's ratios) are not optimal for performance; by adapting the ratio, PLM greatly improves the training process.

\paragraph{Ratio Initialization}
We analyse the importance of the initial ratio, $\bar{r}_{c,1}$, for our algorithm. We run ablations on CIFAR-10 and CIFAR-100 ($\rho=100$), where we initialize the ratio with various values. The results for this ablation can be found in Table \ref{tab:ablationratio}. We find that initializing the ratio with the dataset's ratio ($r_c$) or the mean of the dataset's ratio ($\text{mean}\{r_c\}$) leads to the lowest error. We find that initializing with the maximum or minimum ratio of the dataset tends to lead to poor results. Since the datasets are heavily imbalanced, initializing with either of these ratios leads to too many masked labels which adversely impacts the network's training. We find that initializing the ratio with the dataset's ratio is necessary to achieve good performance on the large-scale dataset MSCOCO. This may be the result of the larger amount of imbalance in MSCOCO or the presence of very difficult classes which have many samples (i.e. a difficult class with many positive samples may require a larger ratio than an easy class with fewer samples).

\begin{table}[t]
\centering
\begin{tabular}{l|c|c}
    \hline
    $\lambda$  & CIFAR-10 & CIFAR-100  \\
    \hline
    1.0 & \textbf{24.86} & 56.92 \\
    0.1 & 25.08 & 57.67 \\
    0.01 & 25.64 & \textbf{55.98} \\
    0.0 (BCE) & 29.43 & 59.88 \\
    \hline
\end{tabular}
\caption{Error rates on CIFAR-10 and CIFAR-100 (both with $\rho=100$). This ablation measures the effect of $\lambda$. For these experiments, the initial ratio is the dataset's ratio $\bar{r}_{c,1}=r_c$. \vspace{-0.2cm}}

\label{tab:ablationlambda}
\end{table}

\begin{table}[t!]
\centering
    \begin{tabular}{l|c|c}
    \hline
    Initial Ratio & CIFAR-10 & CIFAR-100  \\
    \hline
    $\min\{r_c\} $ & 28.82 & 73.46 \\
    $\max\{r_c\} $ & 30.74 & 56.73 \\
    $\text{mean}\{r_c\}$ & 25.65 & 56.08 \\
    $r_c$ & \textbf{24.86} & \textbf{55.98} \\
    \hline
    \end{tabular}
    \caption{Ablations on CIFAR-10  and CIFAR-100 (both with $\rho=100$)  measuring the effect of the initial ratio. $\lambda=1$ and $\lambda=0.01$ for CIFAR-10 and CIFAR-100 respectively. \vspace{-0.3cm}}
    \label{tab:ablationratio}
\end{table}

\section{Discussion and Analysis}
\label{sect:analysis}

\begin{figure*}[t]
\begin{center}
   \includegraphics[width=0.32\linewidth]{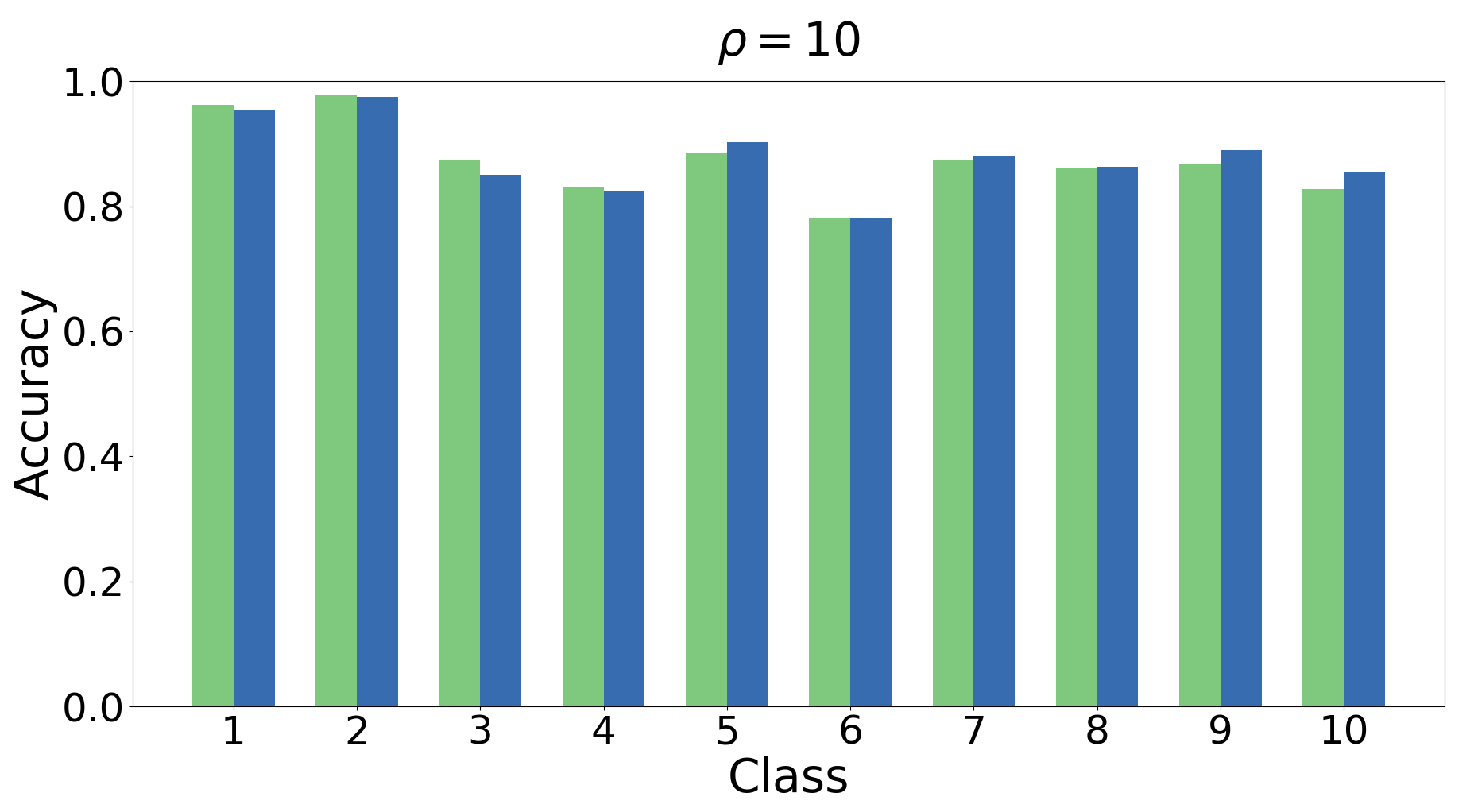}
   \includegraphics[width=0.32\linewidth]{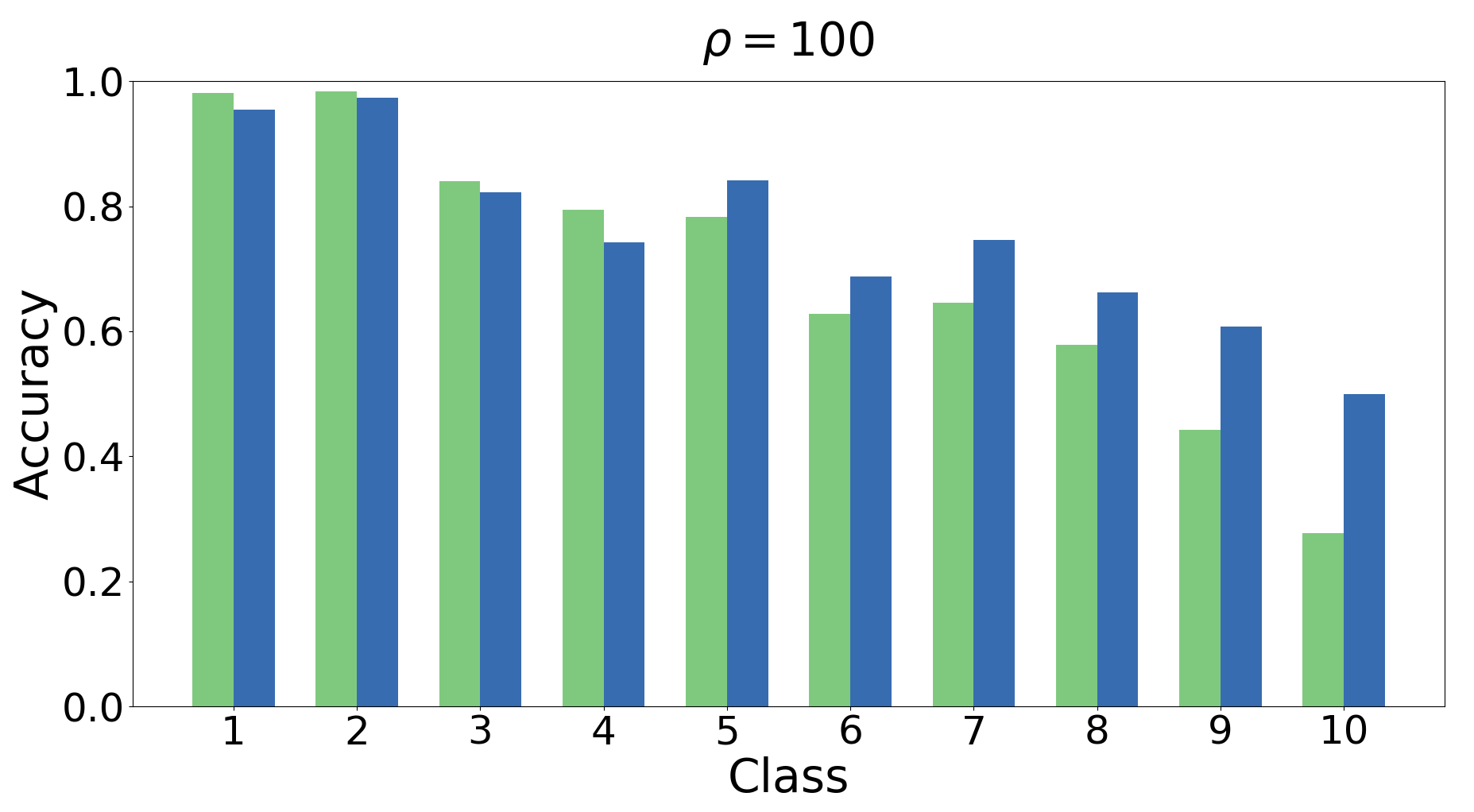}
   \includegraphics[width=0.32\linewidth]{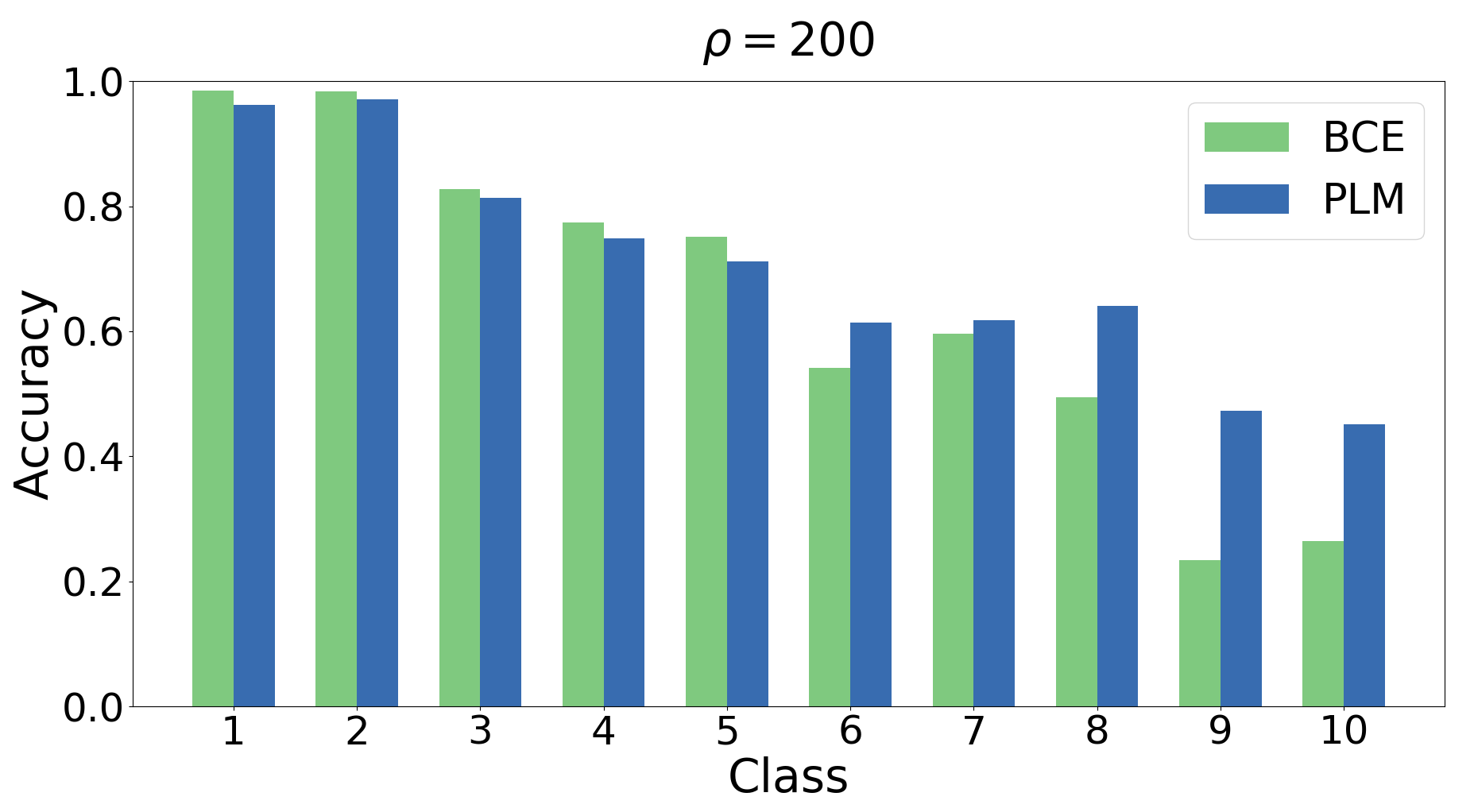}
\end{center}
\vspace{-0.3cm}
   \caption{The test accuracy for each class on CIFAR-10 with varying levels of imbalance. PLM leads to improved generalization on minority classes, especially when there is a large amount of imbalance. \vspace{-0.3cm}}
\label{fig:cifar10class}
\end{figure*}

\begin{figure*}[t!]
\begin{center}
   \includegraphics[width=0.95\linewidth]{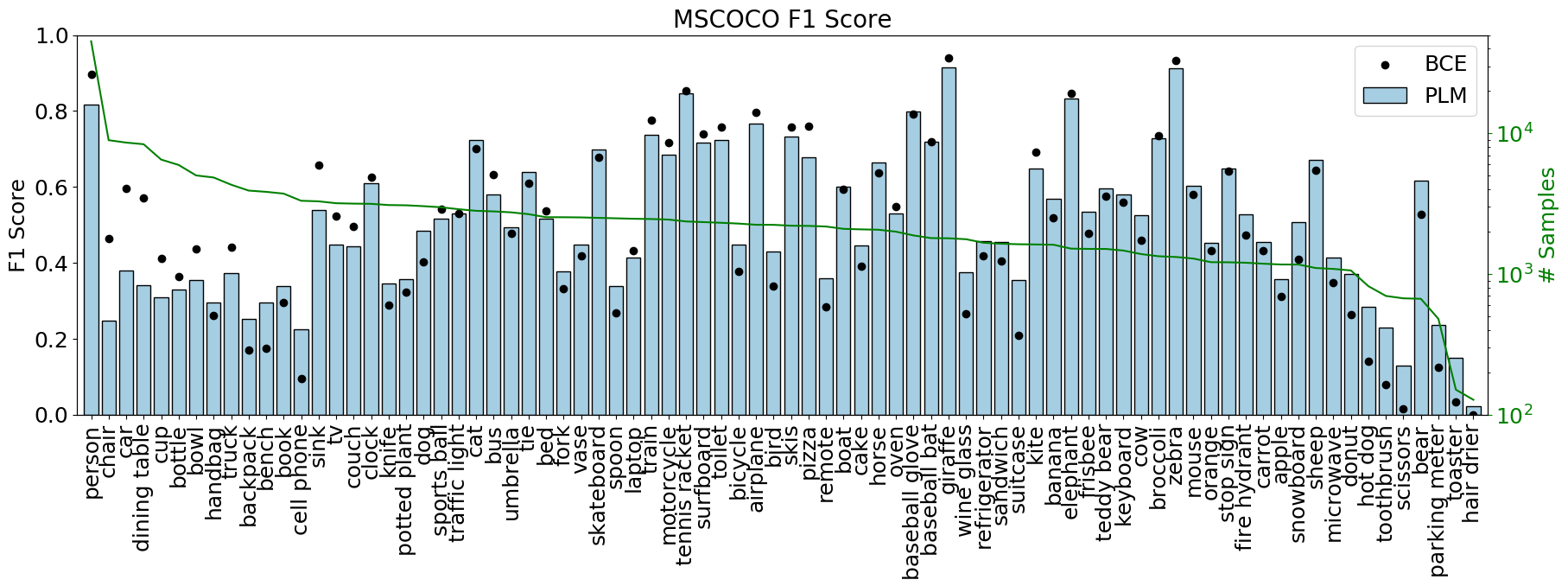}
\end{center}
\vspace{-0.3cm}
   \caption{The class-wise F1 score on the MSCOCO dataset. The classes are ordered based on the number of samples in the training set. The bars represent the results achieved by using PLM, and the points are the results with BCE. Not only does PLM improves results on the 15 most infrequent classes, but also it improves scores on difficult classes, like \textit{backpack}, \textit{bench}, and \textit{cell phone}. \vspace{-0.3cm}}
\label{fig:mscocof1class}
\end{figure*}

\paragraph{Generalization on minority classes}
To perform well on long-tailed datasets, classifiers must generalize well on classes which have few training samples. All of our experiments show that PLM greatly improves performance on minority classes. Figure \ref{fig:cifar10class} displays the class-level accuracy on the CIFAR-10 dataset for various levels of imbalance; as the imbalance factor increases, there is a corresponding relative improvement on tail classes. Similarly, the MSCOCO experiments show an improvement on all 15 of the least frequent classes in terms of F1-score, as seen in Figure \ref{fig:mscocof1class}.

\paragraph{Effects on Precision and Recall}
In imbalanced datasets, classifiers tend to have poor precision on frequent classes and poor recall on infrequent classes. This is due, in part, to the large difference in the ratio of positive to negative samples. By minimizing the KL divergence between the ground-truth and predicted output distributions, PLM effectively balances this ratio, leading to improved performance. On MSCOCO\footnote{We present per-class precision and recall graphs in the Appendix.}, we find that our classifier's precision over the 5 most frequent classes improves by 12.73\% whereas its recall over the 5 least frequent classes improves by 13.04\%. In general, PLM leads to improved recall over most classes, which suggests that it would be best suited for applications which prioritize true positives over false positives. 

\paragraph{Improvements on difficult classes}
Another benefit of using PLM during training is the improvements observed on difficult classes. This behaviour can be seen in Figure \ref{fig:mscocof1class}, with categories \textit{backpack}, \textit{bench}, and \textit{cell phone}. These categories contain many samples, but the network achieves less than $20\%$ F1-score when trained with BCE; when trained with PLM, performance on each of these categories improves by an average of  $11.16\%$. This improvement is caused by the increasing of the ratio $\bar{r}_c$ during training, which leads to masking more negative labels for these classes and an increase in recall.

\paragraph{PLM and Under-sampling} Under-sampling (US) is difficult to apply to the multi-label case due to class co-occurrence. When a head class co-occurs with a tail class, randomly removing samples from the prior can lead to an increase in the level of imbalance (i.e. by removing a sample which contains two classes $i$ and $j$, the imbalance between those classes increases from $n^+_i/n^+_j$ to $(n^+_i-1)/(n^+_j-1)$). Since PLM masks partial labels, it can circumvent this issue by removing the label corresponding to the frequent class while still using the label for the infrequent class. This is evident from the MSCOCO experiments (Table \ref{tab:mscocoresults}) where undersampling to reduced the average f1-score by 2.68\% while PLM led to an increase of 1.42\%. Another difference between the two approaches is that PLM masks both positive and negative labels within a sample, while US only deals with positive labels. This allows PLM to change the ratio of positive and negative labels for a class, without effecting the ratio of other classes.

\section{Related Work}
Many works have studied the problem of long-tailed data imbalance. Extensive overviews of the problem can be found for both classical methods \cite{japkowicz2002class,he2009learning} and convolutional neural networks \cite{buda2018systematic}. 

\paragraph{Resampling Methods}
Resampling techniques either over-sample the infrequent classes or under-sample the frequent classes to reduce the amount of the amount of imbalance seen by the network during training. Several over-sampling approaches \cite{chawla2002smote,han2005borderline,he2008adasyn,mullick2019generative} generate synthetic samples of minority classes which are used alongside real-data to improve classifier performance. Under-sampling \cite{japkowicz2000class,liu2008exploratory} randomly removes some samples from frequent classes during training, and has generally been shown to outperform over-sampling \cite{drummond2003c4}. These techniques have mainly been designed for single-label classification, and do not necessarily extend well to the multi-label case (see Section \ref{sect:analysis}).

\paragraph{Reweighting Methods}
Reweighting methods apply weights to certain classes, or samples, during the objective calculation. The standard approach is to apply class weights based on the inverse number of samples \cite{huang2016learning,wang2017learning,khan2019striking,huang2019deep}, or the square-root of the inverse number of samples \cite{mikolov2013distributed}. Cui \etal \cite{cui2019class} proposed class weights based on the effective number of samples, which improved results on frequent classes. The focal loss \cite{lin2017focal} is another popular technique, which reweighs the loss based on the classifiers' output probabilities. Our method can be viewed as a type of loss reweighting where binary weights are generated stochastically for each sample. Furthermore, PLM is versatile and can be applied on top of existing reweighting and resampling approaches. 

\vspace{-0.2cm}
\paragraph{Additional Long-tail Recognition Methods} Recently, Kang \etal \cite{kang2019decoupling} proposed a method achieving strong long-tailed recognition by decoupling the learned feature representations and the classifier. Liu \etal \cite{liu2019large} propose the task of open long-tailed recognition and present a method which uses dynamic meta-embeddings and modulated attention to transfer knowledge from head to tail classes. Although most long-tail recognition works focus on the single-label classification, a recent work \cite{wu2020distribution} presents a distribution-balanced loss which is effective for multi-label imbalance.

\paragraph{Partial Label Learning}
Our approach takes inspiration from partial label learning, in which learning is performed on data with incomplete, or missing, labels \cite{bucak2011multi,sun2010multi}. Recently, \cite{durand2019learning} proposed an effective method for training a convolutional neural network on partial labels. Whereas the goal of partial label learning is to learn from sparsely annotated samples, our work deals with learning completely annotated imbalanced datasets; our method assumes all labels are complete, and we partially mask some labels to reduce the effect of data imbalance on network training.

\section{Conclusion}

In this work, we propose a general approach to the long-tail data imbalance problem. Our partial label masking algorithm leverages the ratio of positive and negative samples for each class to greatly improve performance on both infrequent and difficult classes. Unlike most data-imbalance approaches, PLM can be successfully applied to the multi-label setting. Furthermore, the method is versatile as it can be used alongside most existing methods for data imbalance. We evaluate PLM on multiple datasets and perform extensive analysis to verify its effectiveness.

\vspace{-0.3cm}
\small\paragraph{Acknowledgments}
This research is based upon work supported by the Office of the Director of National Intelligence(ODNI), Intelligence Advanced Research Projects Activity (IARPA), via IARPA R\&D Contract No. D17PC00345. The views and conclusions contained herein are those of the authors and should
not be interpreted as necessarily representing the official policies or endorsements, either expressed or implied, of the ODNI, IARPA, or the U.S. Government. The U.S. Government is authorized to reproduce and distribute reprints for Governmental purposes not withstanding any copyright annotation thereon.

{\small
\bibliographystyle{ieee_fullname}
\bibliography{main}
}

\clearpage
\newpage
\appendix

\begin{center}
\large
    \textbf{Appendix}
\end{center}

In the appendix we include a description of different multi-label evaluation metrics used (Section \ref{sect:metrics}), a more detailed description of how the discrete output distributions are computed (Section \ref{sec:dists}), and additional details on the imbalanced MultiMNIST dataset (Section \ref{sec:mnistapp}). Furthermore, we include additional results on the MSCOCO dataset (Section \ref{sec:additionalresults}).

\section{Metrics} 
\label{sect:metrics}
In this section, we explain the metrics used in our multi-label classification experiments. In an evaluation set, there are $N$ images with corresponding labels $\{  \textbf{y}^{\left(1\right)}, ..., \textbf{y}^{\left(N\right)}\}$ where $\textbf{y}^{\left(i\right)} = [ y^{\left(i\right)}_1,...,y^{\left(i\right)}_C]$ is the $i^{th}$ binary label vector; note that multiple elements in this label vector may be non-zero in multi-label classification. For each sample, the classifier predicts class probabilities to which a threshold (0.5 in our experiments) is applied to obtain binary prediction vectors $\{  \hat{\textbf{y}}^{\left(1\right)}, ..., \hat{\textbf{y}}^{\left(N\right)}\}$.

\paragraph{Per-class Precision and Recall} The precision and recall of a classifier on a single class $c$ is given by
\begin{equation}
\begin{aligned}
    \text{P}\left(c\right) = \frac{\sum_{i=1}^N \mathbbm{1}\left[y^{\left(i\right)}_c = \hat{y}^{\left(i\right)}_c=1\right]}{\sum_{i=1}^N \mathbbm{1}\left[\hat{y}^{\left(i\right)}_c = 1\right]},\\
    \text{R}\left(c\right) = \frac{\sum_{i=1}^N \mathbbm{1}\left[y^{\left(i\right)}_c = \hat{y}^{\left(i\right)}_c=1\right]}{\sum_{i=1}^N \mathbbm{1}\left[y^{\left(i\right)}_c = 1\right]},\\
\end{aligned}
\end{equation}
respectively, where $\mathbbm{1}\left[...\right]$ is the indicator function. To obtain the reported metric, we average the precision (or recall) over all classes:
\begin{equation}
\begin{aligned}
    \text{Precision} = \frac{1}{C}\sum_{c=1}^C\text{P}\left(c\right),\\
    \text{   Recall} = \frac{1}{C}\sum_{c=1}^C\text{R}\left(c\right).\\
\end{aligned}
\end{equation}
In these metrics, each class is treated equally regardless of the number of samples in the test set.

\paragraph{F1-score} The F1-score is the harmonic mean between precision and recall:
\begin{equation}
    \text{F}_1\left(c\right) =  \frac{2\text{P}\left(c\right)\text{R}\left(c\right)}{\text{P}\left(c\right)+\text{R}\left(c\right)}.
\end{equation}
The final F1-score is averaged over all classes:
\begin{equation}
    \text{F1-score} = \frac{1}{C}\sum_{c=1}^C \text{F}_1\left(c\right).
\end{equation}

\paragraph{0-1 Accuracy} This metric measures how often the network is able to output predictions which exactly match the ground-truth labels. It is computed as follows:
\begin{equation}
    \text{0-1 Accuracy} = \frac{1}{N}\sum_{i=1}^N \mathbbm{1}\left[\textbf{y}^{\left(i\right)} = \hat{\textbf{y}}^{\left(i\right)}\right].
\end{equation}
Since this metric is averaged over all samples, as opposed to all classes, it tends to be biased towards more frequent classes. Therefore, this metric tends to be a poor measure of model performance if there is imbalance present in the test set. 

\section{Output Distributions}
\label{sec:dists}
In this section, we describe in further detail how the discrete distributions $P^+_c$, $\hat{P}^+_c$, $P^-_c$, and $\hat{P}^-_c$ are formed. Given the set of probabilities described in equation 4, we form these discrete distributions through a binning operation. Since the output probabilities are within the range $[0,1]$, we create $\tau$ bins, each of width $1/\tau$. The probabilities are placed within these bins to obtain a histogram, which is then normalized to obtain a discrete probability distribution.

\begin{figure}[t!]
\begin{center}
   \includegraphics[width=0.49\linewidth]{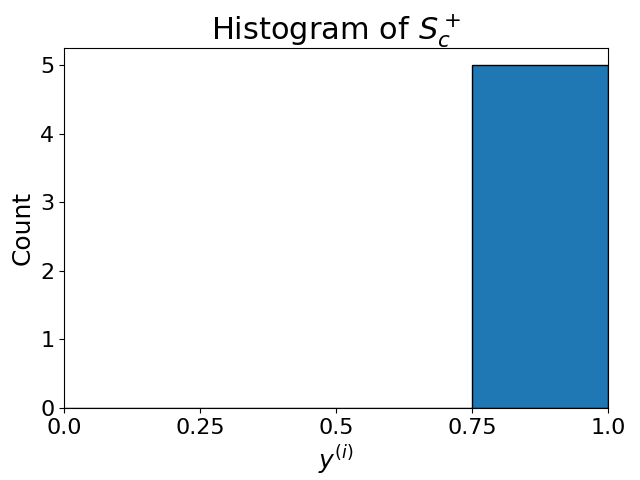}
   \includegraphics[width=0.49\linewidth]{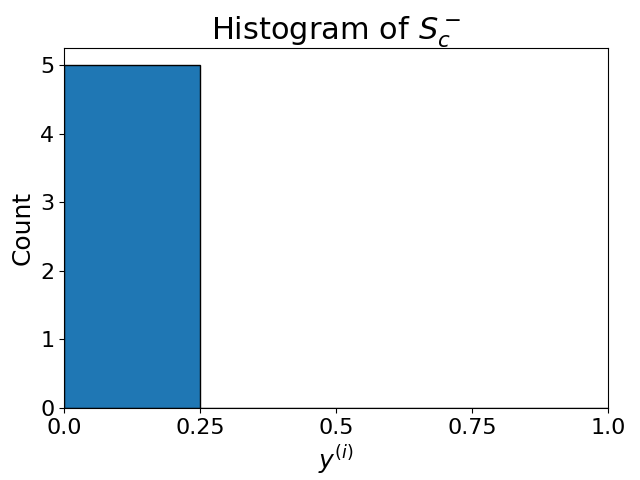}
   \includegraphics[width=0.49\linewidth]{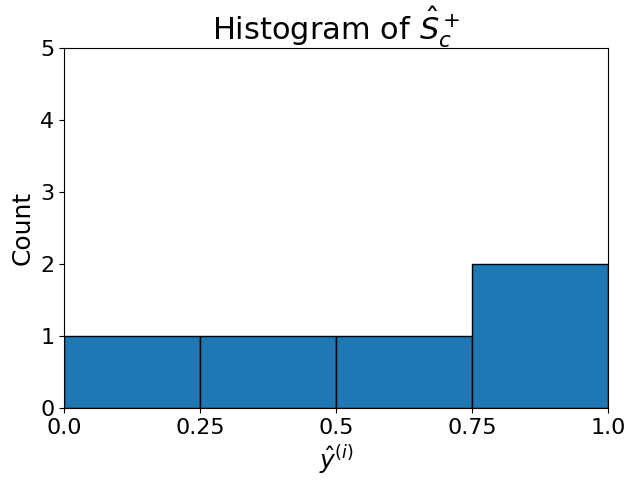}
   \includegraphics[width=0.49\linewidth]{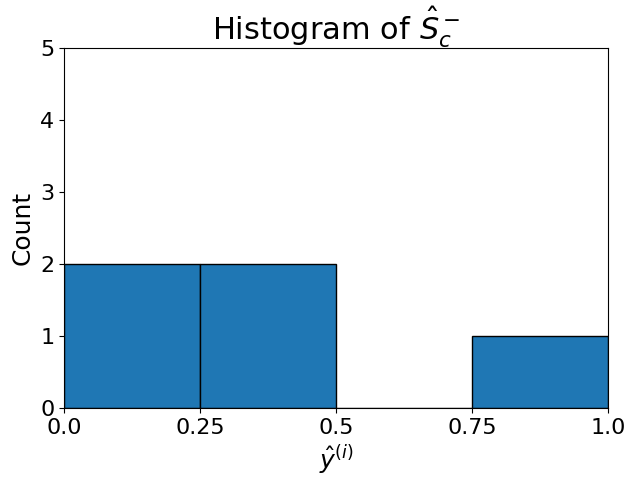}
\end{center}
   \caption{The histograms constructed from the sets of ground-truth labels and predicted probabilities for the toy example described in section \ref{sec:dists}.}
\label{fig:histograms}
\end{figure}

\begin{figure}[t!]
\begin{center}
   \includegraphics[width=0.49\linewidth]{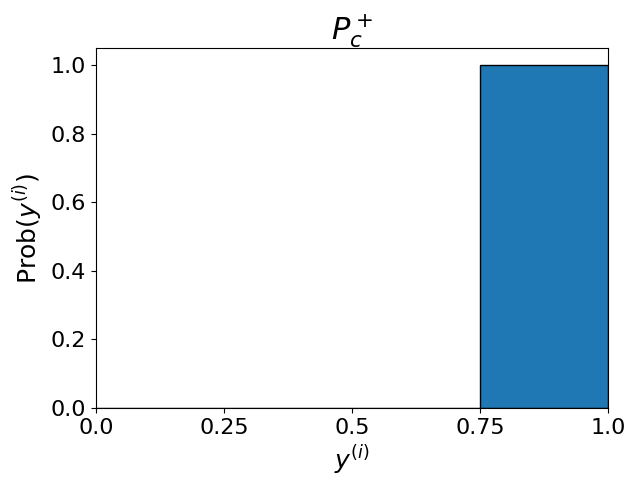}
   \includegraphics[width=0.49\linewidth]{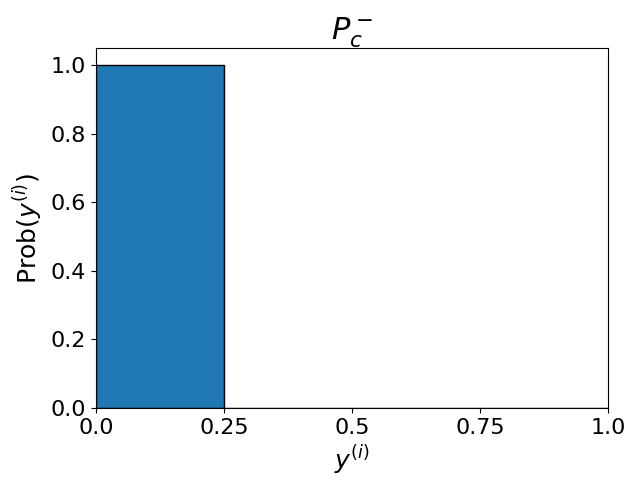}
   \includegraphics[width=0.49\linewidth]{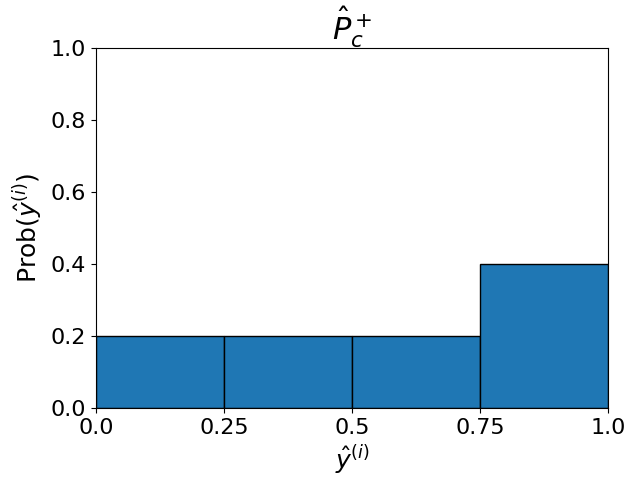}
   \includegraphics[width=0.49\linewidth]{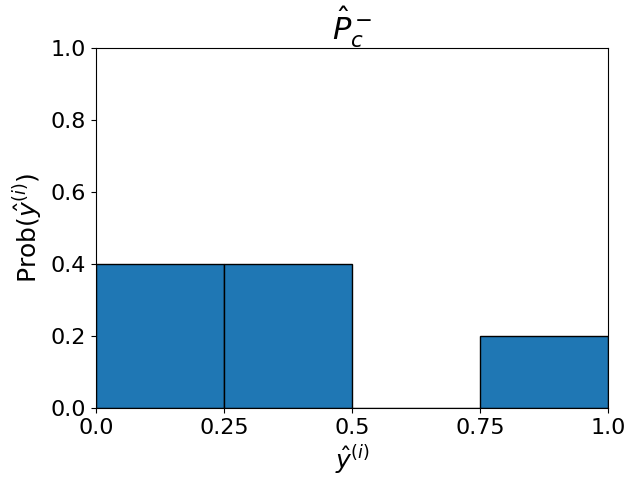}
\end{center}
   \caption{The discrete probability distributions constructed from the sets of ground-truth labels and predicted probabilities for the toy example described in section \ref{sec:dists}.}
\label{fig:probs}
\end{figure}

We present a toy example to describe the process. Suppose there is a single-class dataset with 10 samples and following labels: $\{1, 1, 1, 1, 1, 0, 0, 0, 0, 0 \}$. For these samples, the network predicts probabilities: $\{0.2, 0.6, 0.95, 0.99, 0.45, 0.1, 0.15, 0.8, 0.4, 0.3 \}$. The ground-truth and predicted output sets are as follows: $S^+_c=\{1,1,1,1,1\}$, $S^-_c=\{0,0,0,0,0\}$, $\hat{S}^+_c=\{0.2, 0.6, 0.95, 0.99, 0.45\}$, and $\hat{S}^-_c=\{0.1, 0.15, 0.8, 0.4, 0.3\}$. From these sets we can construct the histogram; the histograms (with $\tau=4$) and their corresponding discrete distributions are shown in figures \ref{fig:histograms} and \ref{fig:probs} respectively. 

\begin{figure*}[t!]
\begin{center}
   \includegraphics[width=0.19\linewidth]{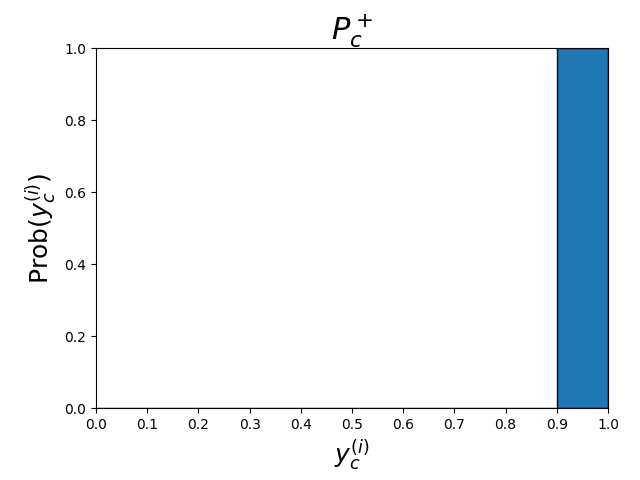}
   \includegraphics[width=0.19\linewidth]{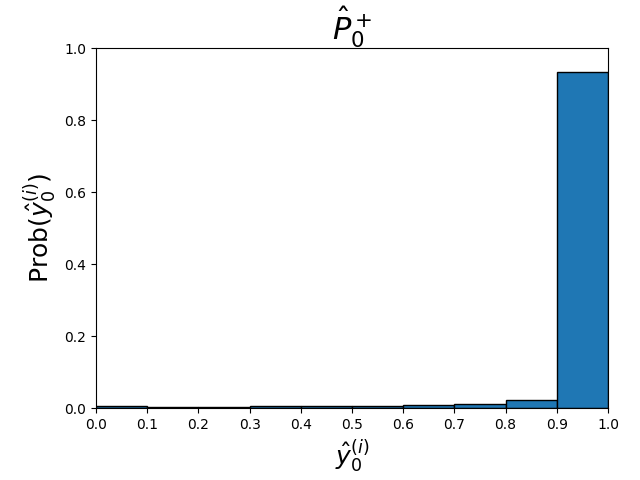}
   \includegraphics[width=0.19\linewidth]{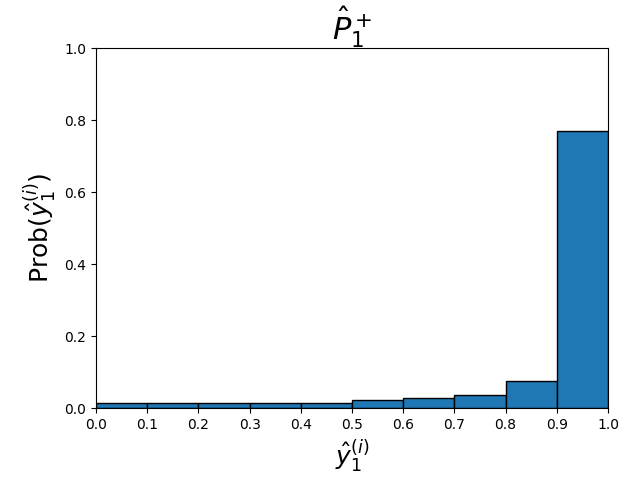}
   \includegraphics[width=0.19\linewidth]{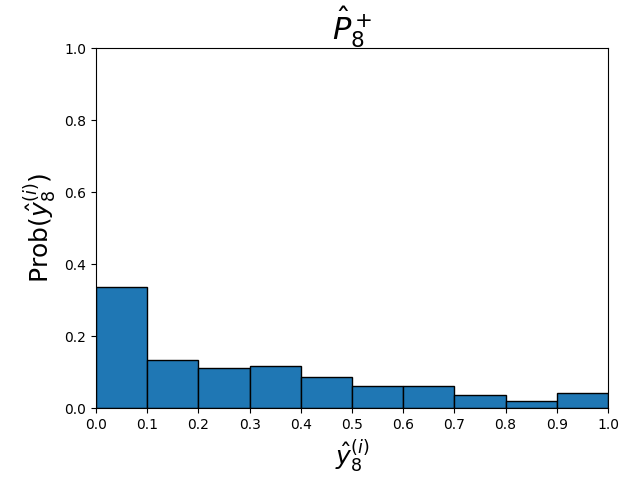}
   \includegraphics[width=0.19\linewidth]{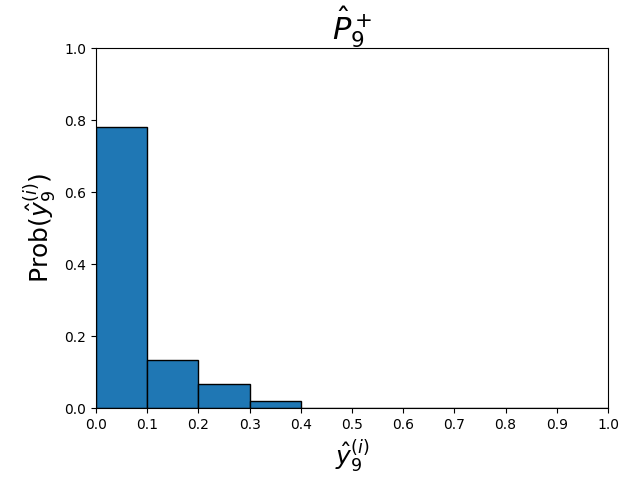}
   
   \includegraphics[width=0.19\linewidth]{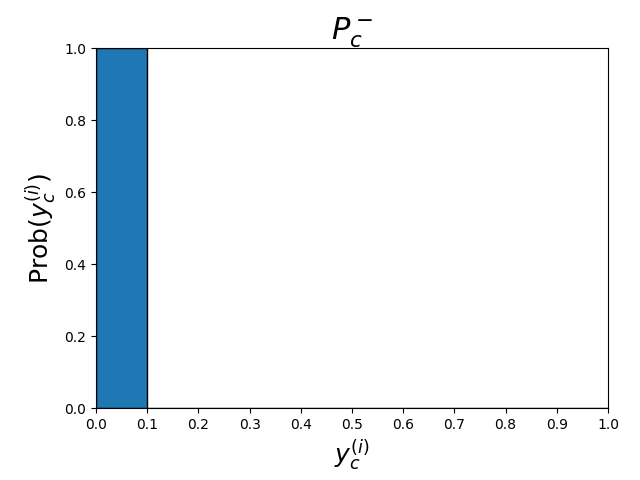}
   \includegraphics[width=0.19\linewidth]{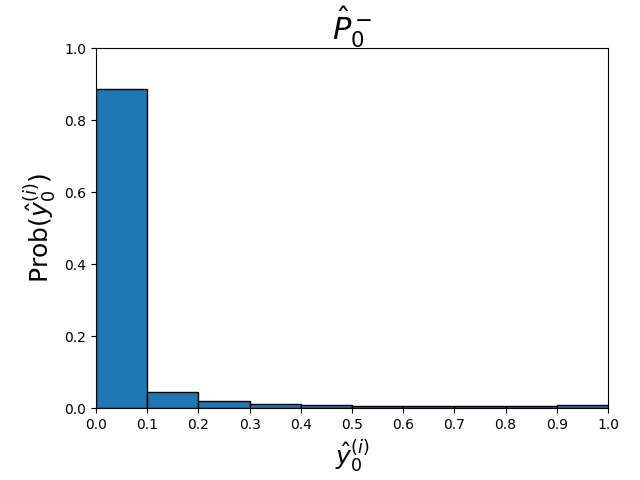}
   \includegraphics[width=0.19\linewidth]{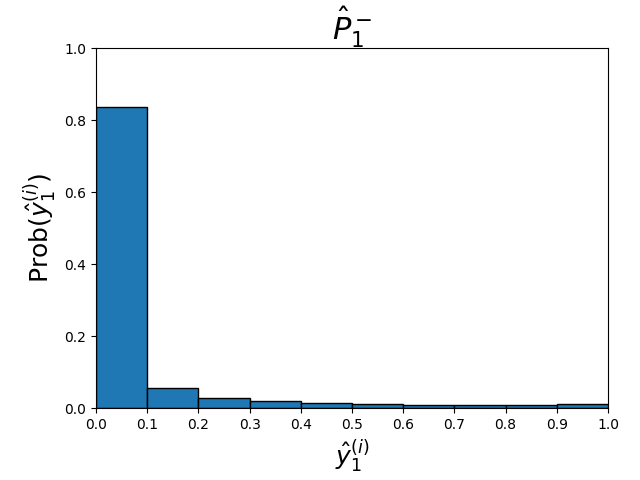}
   \includegraphics[width=0.19\linewidth]{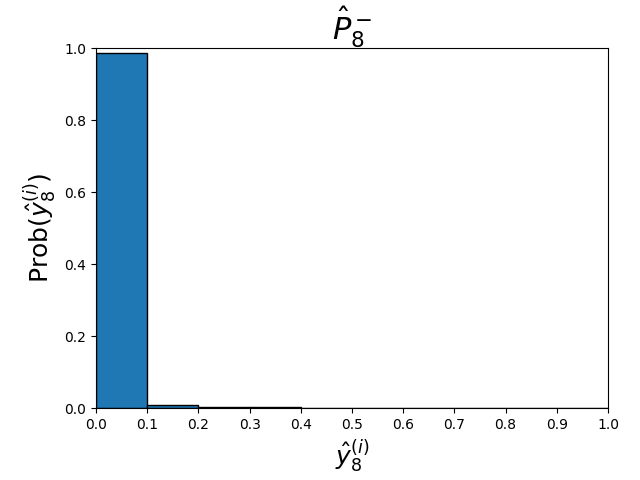}
   \includegraphics[width=0.19\linewidth]{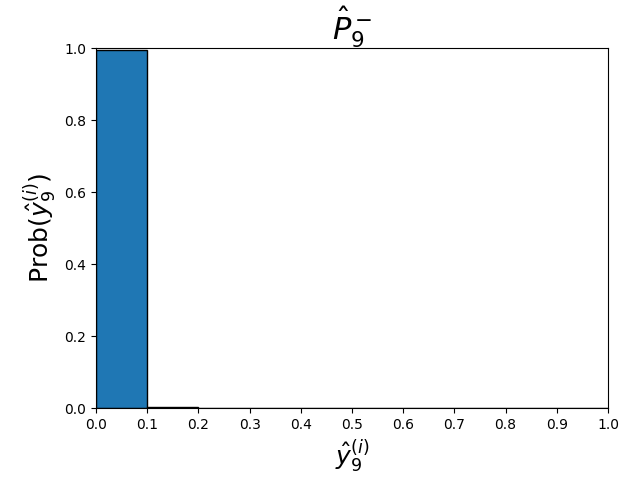}
\end{center}
   \caption{The discrete probability distributions ($\tau=10$) obtained from the MultiMNIST training set. The first column consists of the ground-truth distributions. The distributions on the top are for the positive labels, and the distributions on the bottom are for the negative labels. Distributions for the head classes (columns 2 and 3) tend to be skewed more towards the right (i.e. predictions closer to 1); distributions for the tail classes (columns 4 and 5) tend to be skewed more towards the left (i.e. predictions closer to 0). 
   }
\label{fig:mnistprobs}
\end{figure*}

In general, these distributions tend to be shifted towards the left (i.e. probability outputs closer to zero) for classes with fewer samples and shifted towards the right for classes with many samples. This behaviour can be seen in figure \ref{fig:mnistprobs}, which contains the distributions produced by a classifier when trained on the MultiMNIST dataset for 10 epochs. PLM makes use of this observation to mask individual labels such that predicted probability distributions more resemble to ground-truth distribution.

\section{Imbalanced MultiMNIST Dataset}
\label{sec:mnistapp}

\paragraph{Dataset} We construct the Imbalanced MultiMNIST dataset by superimposing two MNIST \cite{lecun1998mnist} digits into a single image. We begin by sampling the MNIST training set similarly to how \cite{cui2019class} create the Imbalanced CIFAR dataset with an imbalance of $\rho=100$. These images are padded to become $32\times32$ and shifted randomly by $-6$ to $6$ pixels, both vertically and horizontally. Then, each image is randomly paired with another digit in the training set, which is also padded and shifted randomly. These two digits are superimposed upon each other to create a $32\times32$ image with two digits. Due to cooccurrence of the same digit, within a sample, the final imbalance of the dataset becomes $90.33$. 

For evaluation, we select all samples in the MNIST test set and perform the same padding and shifting operations. For each digit, we select other random test digits that have different labels, which are superimposed to make 9 samples with two digits each. This results in 90000 test samples. The test set is roughly balanced, with all classes having a similar number of samples. The number of samples for each class can be found in Table \ref{tab:multimnistimbalance}.

\begin{table}[t]
    \centering
    \begin{tabular}{l|c|c}
        \hline
        Digit  & Train & Test  \\
        \hline
        0 & 9485 & 17840\\
        1 & 6167 & 19080\\
        2 & 3967 & 18256\\
        3 & 2476 & 18080\\
        4 & 1508 & 17856\\
        5 & 875 & 17136\\
        6 & 538 & 17664\\
        7 & 331 & 18224\\
        8 & 164 & 17792\\
        9 & 105 & 18072\\
        \hline
        Total & 14694 & 90000\\
        \hline
    \end{tabular}
    
    \caption{The number of samples for each digit in the Imbalanced MultiMNIST dataset.}
    \label{tab:multimnistimbalance}
\end{table}

\section{Additional Results}
\label{sec:additionalresults}

\begin{figure*}[t!]
\begin{center}
   \includegraphics[width=\linewidth]{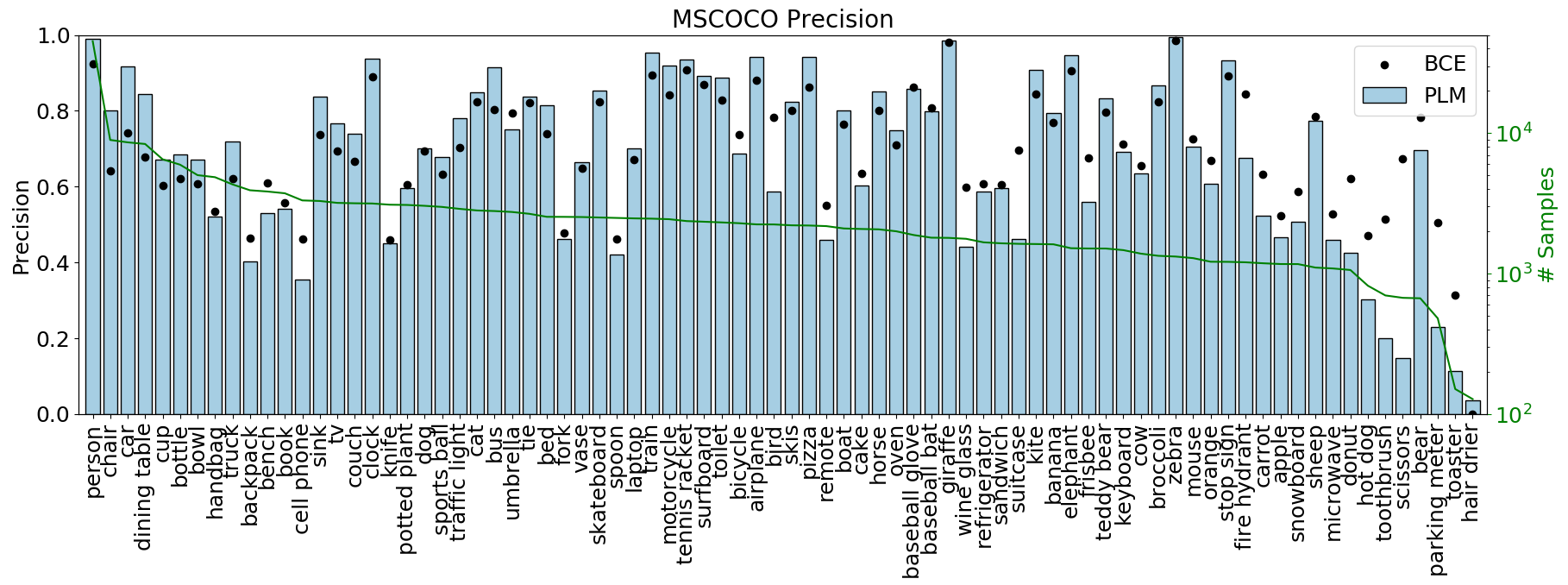}
   \includegraphics[width=\linewidth]{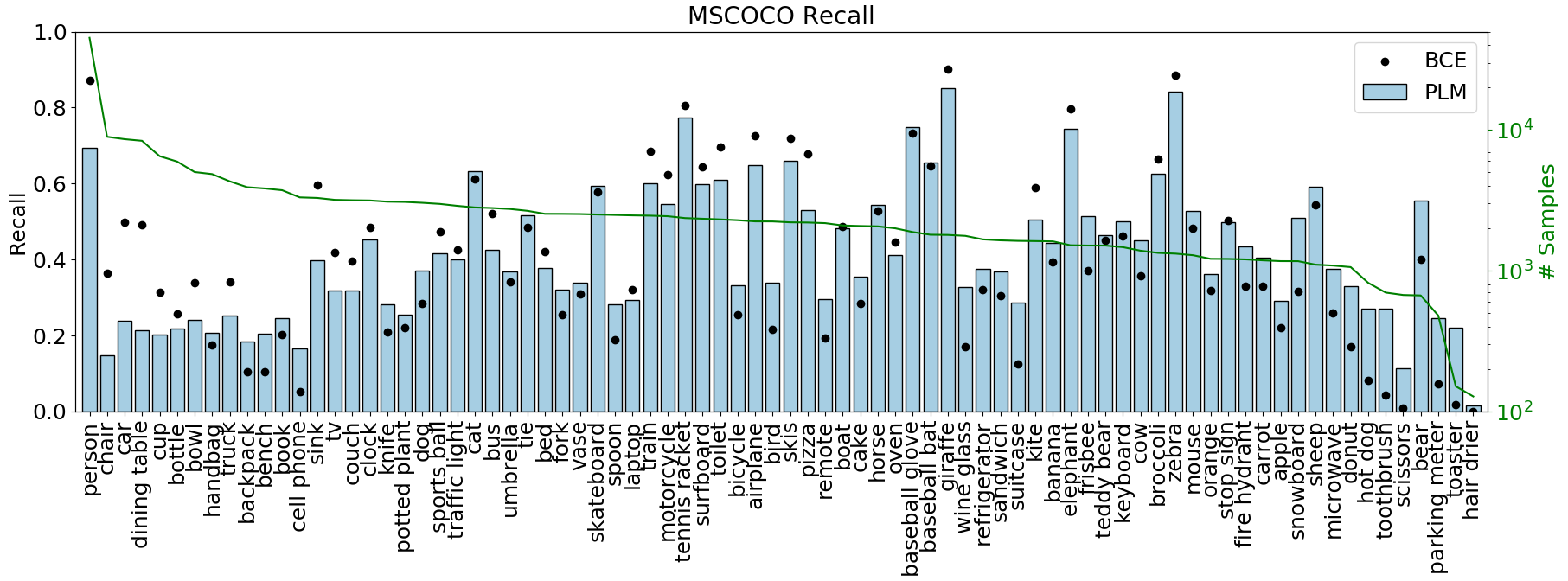}
\end{center}
   \caption{The class-wise precision and recall scores on the MSCOCO dataset. The classes are ordered based on the number of samples in the training set. The bars represent the results achieved by using the PLM method, and the points are the results when using BCE. }
\label{fig:mscocoprclass}
\end{figure*}

\paragraph{Additional MSCOCO Results} We include more results on the MSCOCO dataset. Figure \ref{fig:mscocoprclass} contains the class-wise precision and recall scores on the MSCOCO dataset. We find that using PLM leads to an large improvement in recall for the tail classes, and a slight improvement in precision for the head classes (7.86\% improvement on the 10 most frequent classes). This is in line with the observation that PLM reduces over-prediction on the head classes while reducing under-prediction on the tail classes (as well as some difficult classes).

\end{document}